%% file: 0-main.tex
\documentclass[bst/sn-mathphys]{sn-jnl}

\jyear{2021}%

\usepackage{algorithm}

\theoremstyle{thmstyleone}%
\theoremstyle{thmstyletwo}%

\theoremstyle{thmstylethree}%
\newtheorem{definition}{Definition}%

\usepackage{balance} 
\usepackage{perpage}
\MakePerPage{footnote}

\usepackage{graphicx}
\usepackage[caption=false,font=normalsize,labelfont=sf,textfont=sf]{subfig}
\usepackage{caption}
\usepackage{hyperref}
\usepackage{amssymb}

\usepackage{amsmath,amssymb,amsthm}

\usepackage{booktabs} 
\usepackage{multirow}
\usepackage{multicol}
\usepackage{bm} 
\usepackage{stfloats}
\usepackage{enumitem}
\usepackage{float}
\usepackage{booktabs}

\usepackage{array} 

\newcommand{\revise}[1]{{\color{black}#1}}

\begin{document}

\title[Learning Top-K Subtask Planning Tree]{Learning Top-K Subtask Planning Tree based on Discriminative Representation Pretraining for Decision-Making}


\author[1,2]{\fnm{Jingqing} \sur{Ruan}}

\author[1,3]{\fnm{Kaishen} \sur{Wang}}

\author[1,2]{\fnm{Qingyang} \sur{Zhang}}

\author[1,3]{\fnm{Dengpeng} \sur{Xing}$^\dagger$}

\author[1,3]{\fnm{Bo} \sur{Xu}$^\dagger$}

\affil[1]{\orgdiv{Institute of Automation}, \orgname{Chinese Academy of Sciences}, \orgaddress{\city{Beijing} \postcode{100190},  \country{China}}}

\affil[2]{\orgdiv{School of Future Technology}, \orgname{University of Chinese Academy of Sciences}, \orgaddress{\city{Beijing} \postcode{100049},  \country{China}}}

\affil[3]{\orgdiv{School of Artificial Intelligence}, \orgname{University of Chinese Academy of Sciences}, \orgaddress{\city{Beijing} \postcode{100049},  \country{China}}}


\abstract{Decomposing complex real-world tasks into simpler subtasks and devising a subtask execution plan is critical for humans to achieve effective decision-making.
However, replicating this process remains challenging for AI agents and naturally raises two questions: (1) How to extract discriminative knowledge representation from priors? (2) How to develop a rational plan to decompose complex problems?
\revise{
To address these issues, we introduce a groundbreaking framework that incorporates two main contributions. First, our multiple-encoder and individual-predictor regime goes beyond traditional architectures to extract nuanced task-specific dynamics from datasets, enriching the feature space for subtasks. Second, we innovate in planning by introducing a top-$K$ subtask planning tree generated through an attention mechanism, which allows for dynamic adaptability and forward-looking decision-making. Our framework is empirically validated against challenging benchmarks BabyAI including multiple combinatorially rich synthetic tasks (e.g., GoToSeq, SynthSeq, BossLevel), where it not only outperforms competitive baselines but also demonstrates superior adaptability and effectiveness in complex task decomposition.
}

}





\maketitle

\def\thefootnote{$^\dagger$}\footnotetext{Corresponding authors.}\def\thefootnote{\arabic{footnote}}

\input{1-intro}

\input{2-related}

\input{3-pre}
\input{4-method}

\input{5-experiments}
\input{6-con}

\input{7-ack}

\clearpage

\bibliographystyle{bst/sn-mathphys}
\bibliography{sample}
%

\input{7-append}

\clearpage

\begin{figure}[h]%
\centering
\includegraphics[width=0.4\textwidth]{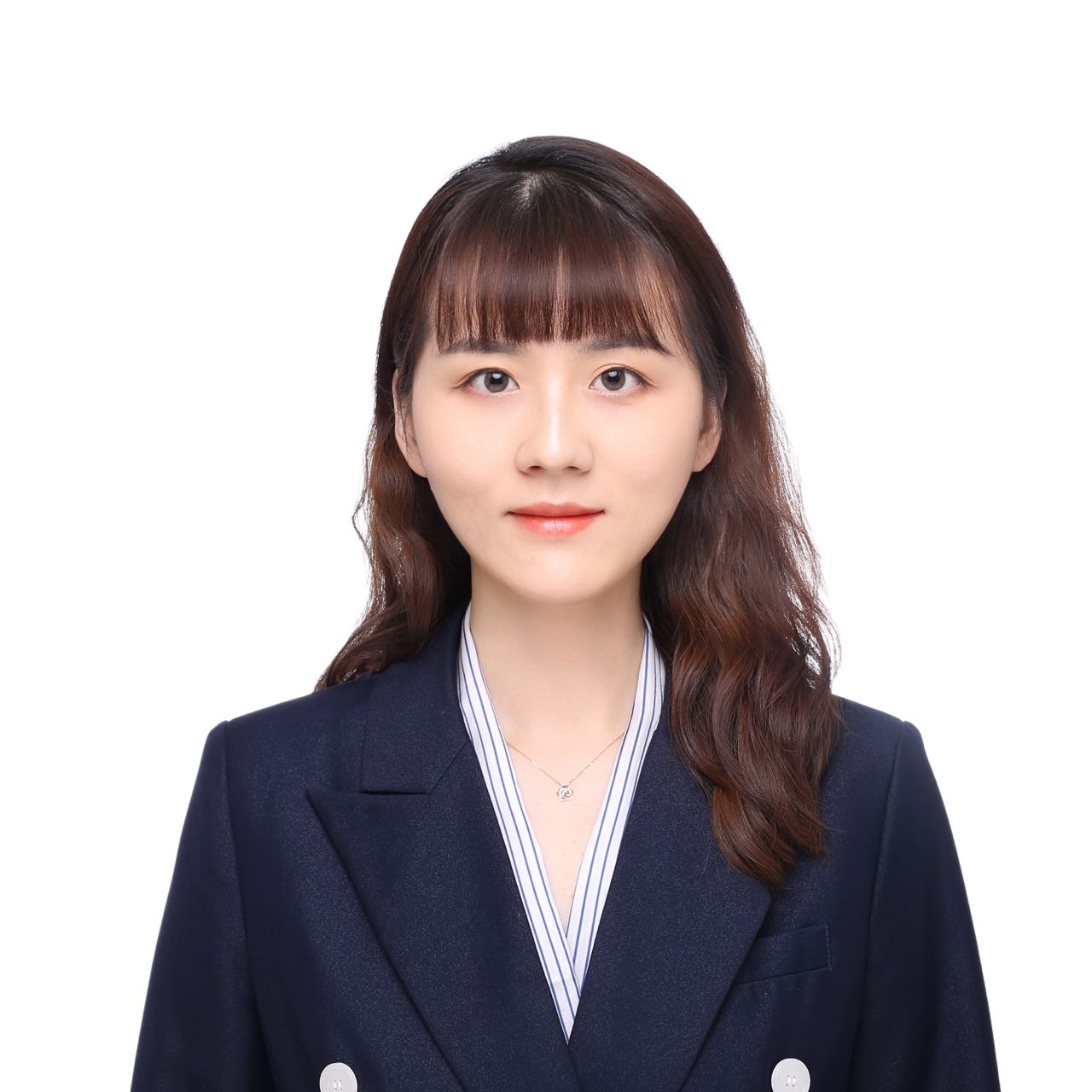}
\end{figure}
\noindent{\bf Jingqing Ruan }\quad received a B.Sc. degree in software engineering from North China Electric Power University, China, in 2019. She is currently a Ph.D. candidate in pattern recognition and intelligent systems at the Institute of Automation,
Chinese Academy of Sciences and the University of Chinese Academy of Sciences, China. 

Her research focuses on multi-agent reinforcement learning, multi-agent planning on game AI, and graph-based multi-agent coordination. 

E-mail: ruanjingqing2019@ia.ac.cn

ORCID ID: 0000-0002-4857-9053

\begin{figure}[h]%
\centering
\includegraphics[width=0.3\textwidth]{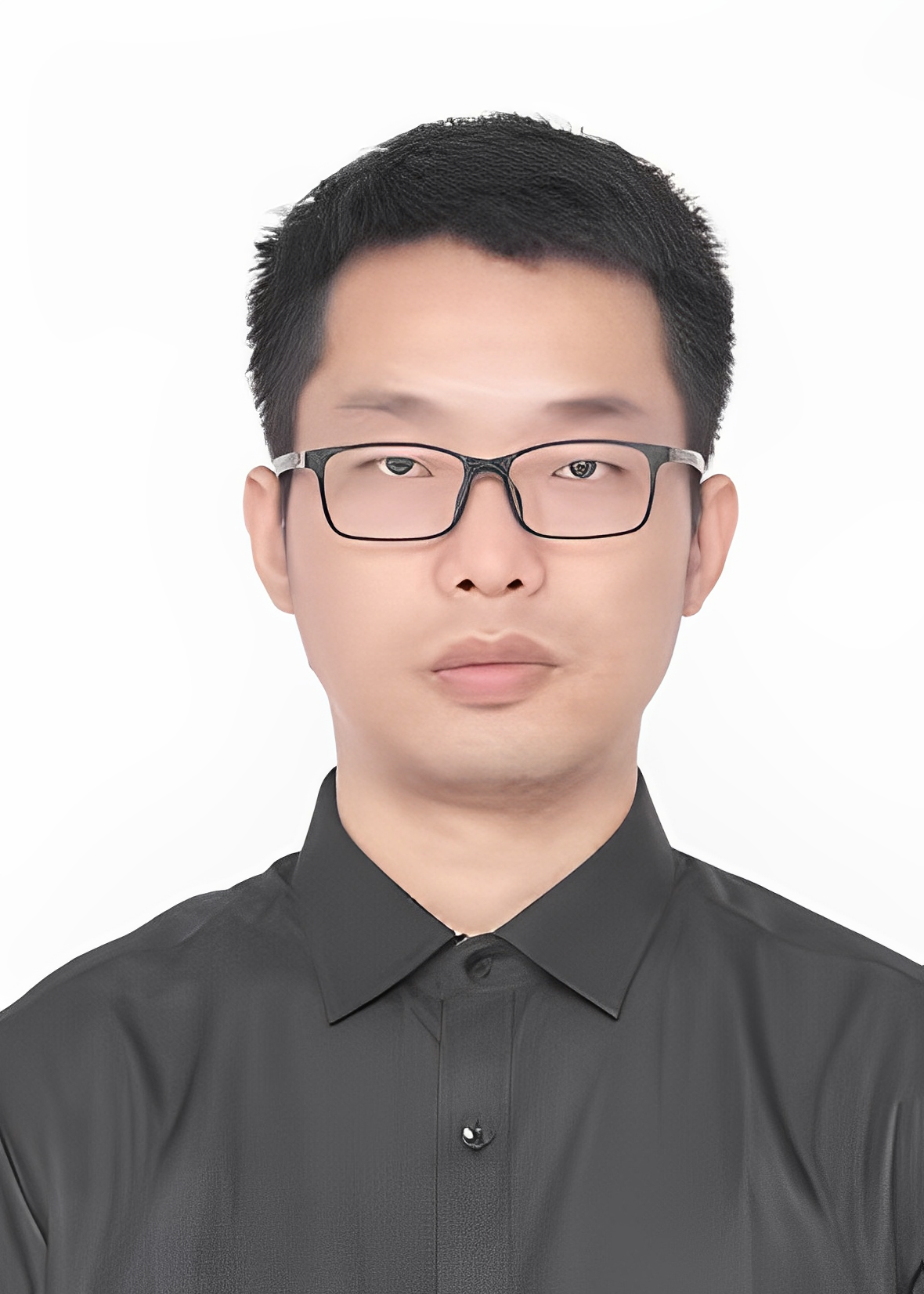}
\end{figure}
\noindent{\bf Kaishen Wang }\quad received a B.Sc. degree in electrical engineering and automation from China University of Mining and Technology, China, in 2015.
He is currently an M.S. candidate in computer science and technology at the Institute of Automation,
Chinese Academy of Sciences and the University of Chinese Academy of Sciences, China.

His research interests include hierarchical reinforcement learning and multi-agent reinforcement learning.

E-mail: wangkaishen2021@ia.ac.cn

ORCID ID: 0000-0002-2787-5873


\begin{figure}[h]%
\centering
\includegraphics[width=0.3\textwidth]{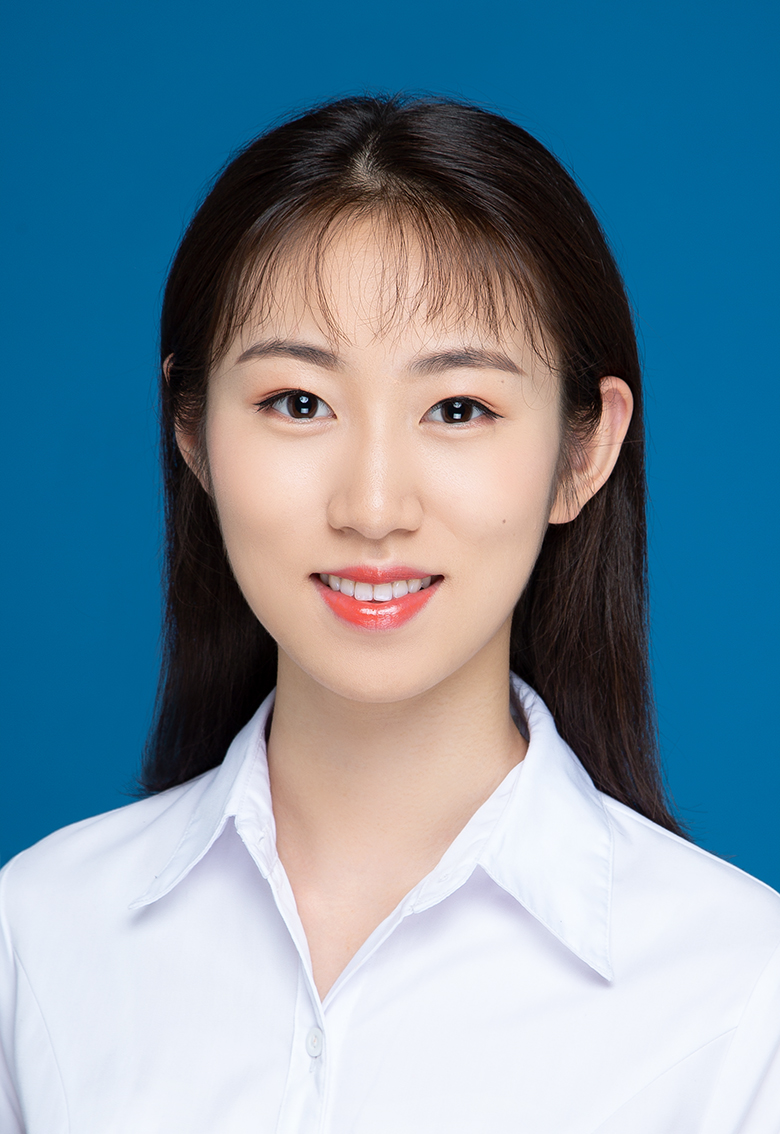}
\end{figure}

\noindent{\bf Qingyang Zhang }\quad received a B.Sc. degree in communication engineering from Shandong University, China, in 2019.
She is currently a Ph.D. candidate in pattern recognition and intelligent systems at the Institute of Automation,
Chinese Academy of Sciences and the University of Chinese Academy of Sciences, China.

Her research interests include hierarchical reinforcement learning, contrastive learning, representation learning, and multi-agent reinforcement learning.

E-mail: zhangqingyang2019@ia.ac.cn

ORCID ID: 0000-0001-5387-9942

\begin{figure}[h]%
\centering
\includegraphics[width=0.3\textwidth]{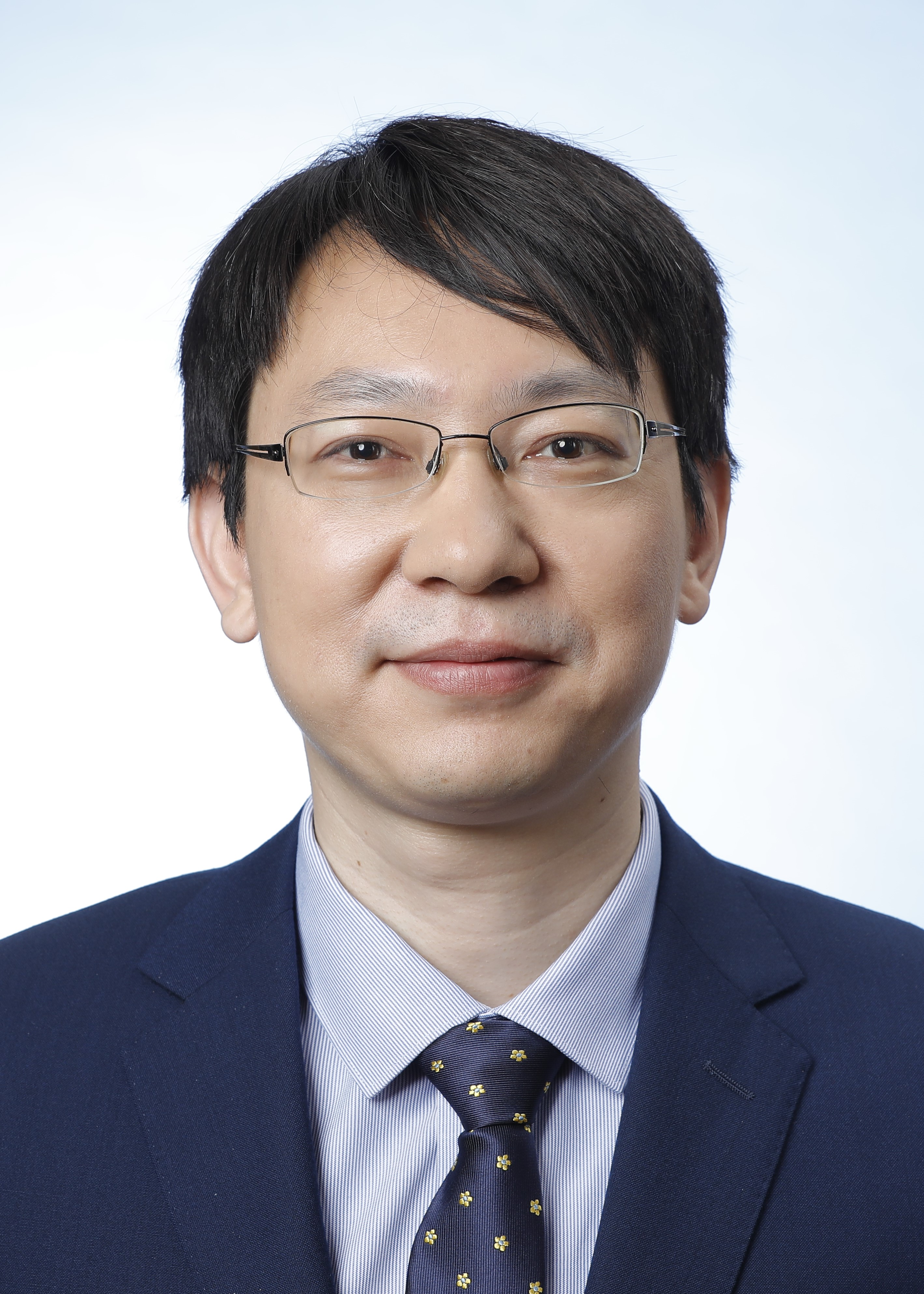}
\end{figure}

\noindent{\bf Dengpeng Xing}\quad obtained his B.Sc. degree in Mechanical Electronics and M.Sc. degree in Mechanical Manufacturing and Automation from Tianjin University, China, in 2002 and 2006, respectively. He earned his Ph.D. degree in Control Science and Engineering from Shanghai Jiao Tong University, China, in 2010. Currently, he holds the position of Professor at the National Key Laboratory for Multi-modal Artificial Intelligence Systems, Institute of Automation, Chinese Academy of Sciences, China.

Dr. Xing's research focuses on robot control, reinforcement learning, and causal learning.

E-mail: dengpeng.xing@ia.ac.cn (Corresponding author)

ORCID ID: 0000-0002-8251-9118





\begin{figure}[h]%
\centering
\includegraphics[width=0.3\textwidth]{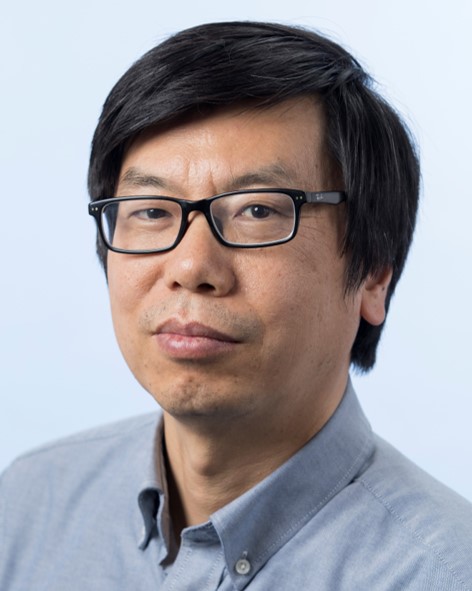}
\end{figure}
\noindent{\bf Bo Xu }\quad is a professor, the director of the Institute of Automation Chinese Academy of Sciences, and deputy director of the Center for Excellence in Brain Science and Intelligence Technology, Chinese Academy of Sciences. 

His main research interests include brain-inspired intelligence, brain-inspired cognitive models, natural language processing and understanding, and brain-inspired robotics.

E-mail: xubo@ia.ac.cn (Corresponding author)

ORCID ID:  0000-0002-1111-1529

\end{document}

%% file: 1-intro.tex
\section{Introduction}

\begin{figure}[ht!]
    \centering
\includegraphics[width=0.7\linewidth]{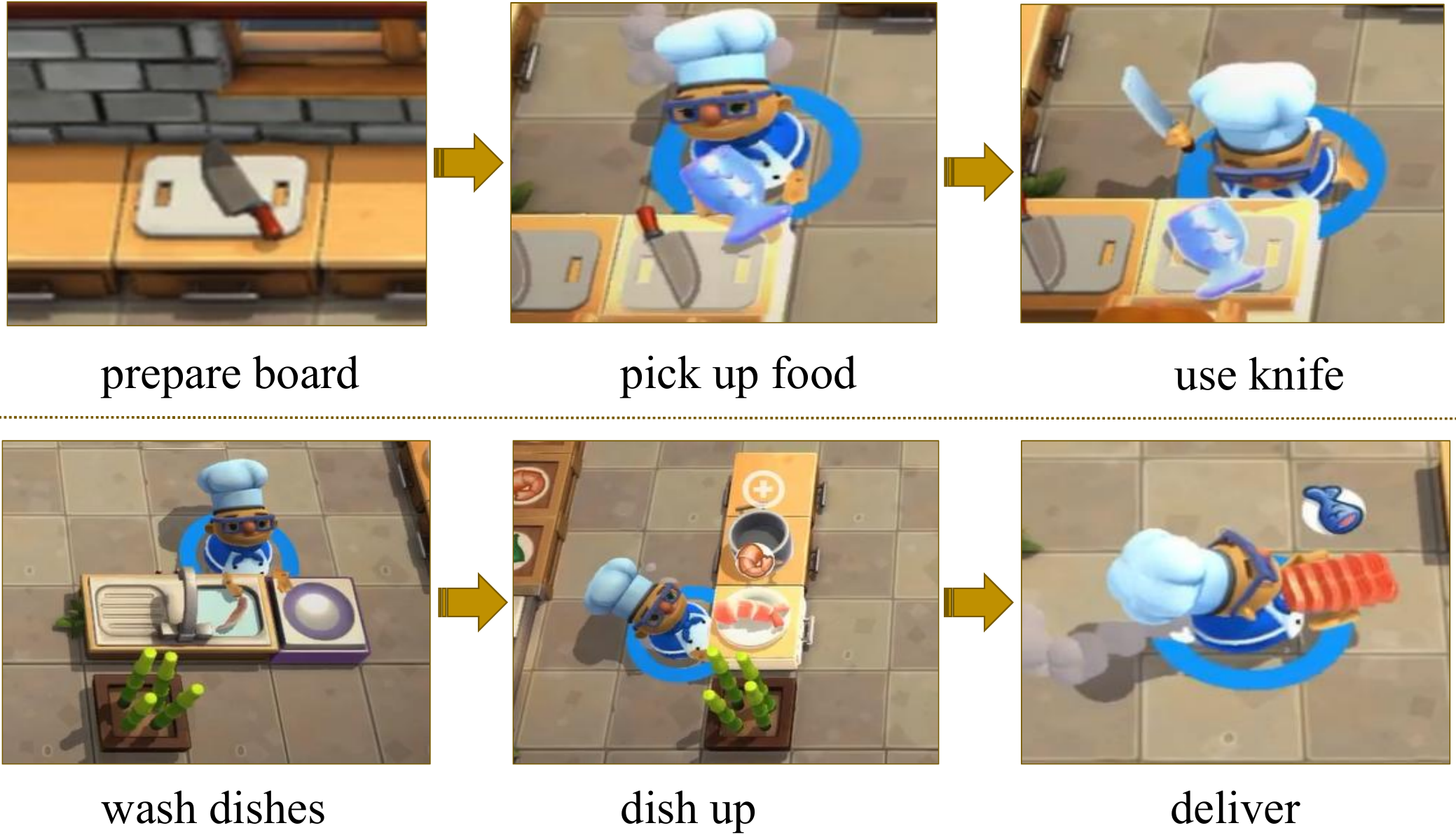}
\caption{\revise{Taking two scenarios in Overcooked AI~\cite{carroll2019utility} as examples, the top denotes cutting food, and the bottom is serving customers. Cutting food should take command of solving a sequence of subtasks: prepare the chopping board, pick up an onion, and use a knife to cut. The execution of serving customers should be performed: wash dishes, dish up, and deliver.}}
\label{fig:intro}
\end{figure}

%
Decomposing complex problems into smaller, and more manageable subtasks with some priors and then solving them step by step is one of the critical human abilities.
In contrast, reinforcement learning (RL)~\cite{schulman2017ppo,haarnoja2018sac,konda1999actor_critic,dang2022dynamic} agents typically learn from scratch and need to collect a large amount of experience, which can be costly, particularly in real-world scenarios. 
To emulate the efficient decision-making process of humans, one solution is to leverage subtask priors to extract useful knowledge representations and transduce them into decision conditions to assist policy learning.
Many real-world challenging problems with different task structures can be solved with the subtask-conditioned reinforcement learning (ScRL) regime, such as robot skill learning~\cite{kase2020transferable,mulling2013learning,cambon2009hybrid,li2021skill,wurman2008coordinating}, navigation~\cite{zhu2017target,zhu2021deep,quan2020novel,gupta2022intention}, and cooking~\cite{carroll2019utility,knott2021evaluating,sarkar2022pantheonrl}. 
ScRL endows an agent with the learned knowledge to perform corresponding subtasks, and the subtask execution path should be planned in a rational manner. Ideally, the agent should master some basic competencies to finish some subtasks and composite the existing subtasks to find an optimal decision path to solve previously unseen and complicated problems. Taking the task of cutting onions for example, consists of multiple interdependent subtasks, including preparing a chopping board, procuring an onion, and utilizing a knife. \revise{For clarity, we visualize two different task structures by recombining subtask sequences, as shown in Fig.~\ref{fig:intro}.} It is crucial to integrate the subtasks in a sequential manner to accomplish the ultimate goal of onion cutting.

However, ScRL faces two major challenges. The first challenge involves the construction of proper knowledge representations for subtasks, which can facilitate the transfer to more complex tasks. The content and structure of these representations are critical to decision-making, as they offer functional descriptions of different subtasks leading to the completion of the final goal. \revise{Addressing this challenge requires a nuanced understanding of the dynamics of each subtask, which we propose can be learned from their trajectories. To operationalize this, we introduce a novel learning regime based on multiple encoders and an individual predictor aimed at extracting distinguishable subtask representations that encapsulate these dynamics along with other task-essential properties. This framework utilizes contrastive learning and model-based dynamic prediction to capture these features effectively. Importantly, each encoder is designated to a specific subtask, minimizing interference from other subtasks and enhancing the quality of the extracted representations.}

Another key challenge in ScRL is how to leverage learned knowledge to improve policy training. While prior works~\cite{netanyahu2022discovering,kwan2023survey,jang2022bc,ryu2022remax} have primarily focused on zero-shot generalization to unseen tasks of the similar complexity, little attention has been paid to the recombination of prior knowledge to plan more intricate tasks that can guide policy learning. \revise{This narrow focus limits the agent's ability to adapt to more complex or unfamiliar scenarios. To address this limitation, we introduce a novel construction algorithm that generates a top-$K$ subtask planning tree, customizing the execution plan for each subtask. This approach not only tailors the learning process to the specific needs of the agent but also allows for the incorporation of learned knowledge into the decision-making policy. Unlike traditional methods, the top-$K$ subtask planning tree can be expanded to any width and depth, providing a flexible and scalable solution for complex tasks. 
Our top-$K$ subtask planning tree facilitates a more comprehensive view by enabling the agent to look ahead and consider a range of possible future subtasks and their implications. By doing so, the agent can more accurately weigh the immediate rewards against potential future gains or losses. This long-term outlook minimizes the chances of the agent getting stuck in locally optimal solutions that are suboptimal in the broader context. In essence, long-term planning adds a layer of foresight that contributes to optimizing the overall decision-making process, making the agent better equipped to handle complex tasks.
}

In summary, the contributions of this paper can be summarized in threefold.
\begin{itemize}
    \item We introduce a multiple-encoder and individual-predictor learning regime to learn distinguishable subtask representations with task-essential properties, which enables the agent to acquire generalized subtask knowledge.
    \item We propose to construct a top-$K$ subtask planning tree to customize subtask execution plans for learning with subtask-conditioned reinforcement learning.
    \item Empirical evaluations on several challenging navigation tasks (e.g., GoToSeq, SynthSeq, BossLevel) show that our method can achieve superior performance and obtain meaningful results consistent with intuitive expectations.
\end{itemize}

%% file: 2-related.tex
\section{Related Work}

{\textbf{Task representation.}}
One line of research, including HiP-BMDP~\cite{zhang2020learning}, MATE~\cite{schafer2022learning}, 
MFQI~\cite{d2020mfqi}, CARE~\cite{sodhani2021care}, CARL~\cite{benjamins2021carl}, etc, are designed to achieve fast adaptation by aggregating experience into a latent representation on which policy is conditioned to solve the zero/few-shot generalization in multi-task RL.
Another line, such as CORRO~\cite{yuan2022corro}, MerPO~\cite{lin2021MerPO},  FOCAL~\cite{li2020focal}, \revise{CMT~\cite{lin2022contextual}, HS-OMRL~\cite{zhao2023context}}, etc, mainly focus on learning the task/context embeddings with fully offline meta-RL to mitigate the distribution mismatch of behavior policies in training and testing. 
Among all the aforementioned approaches, this paper is the most related to the latter. Part of our work is devoted to finding the proper subtask representations using offline datasets to assist online policy learning.

{\textbf{Task planning.}}
Task planning essentially breaks down the final goals into a sequence of subtasks.
Based on the style of planning, we classify the existing approaches into \revise{four} categories: planning with priors, planning on latent space, and planning with implicit knowledge.
First, planning with priors will generate one or more task execution sequences according to the current situation and the subtask function prior, such as~\cite{kuffner2000rrt,liang2022search,garrett2021integrated,okudo2021online}, 
but is limited by subtasks divided by human labor.
Second, planning on latent space~\cite{jurgenson2020sub,pertsch2020long,zhang2021world,ao2021co,zhang2018learning} endows agents with the ability to reason over an imagined long-horizon problem, but it is fundamentally bottlenecked by the accuracy of the learned dynamics model and the horizon of a task.
Third, some works such as~\cite{kaelbling2017learning,yang2022ldsa,yang2020multi,ruan2023learning}, utilize the modular combination for implicit knowledge injection to achieve indirect task guidance and planning, which lack explainability.
\revise{Finally, recent developments in the field have seen the emergence of work such as TPTU~\cite{ruan2023tptu}, ProgPrompt~\cite{singh2023progprompt}, Plan4mc~\cite{yuan2023plan4mc}, and Voyager~\cite{wang2023voyager}, which leverage large language models (LLMs) for task planning. However, a significant drawback of using LLMs is their substantial computational cost.}
Our work aims to obtain proper knowledge representation extracted from the subtask and generate a rational subtask tree including one or more task execution sequences.

{\textbf{Subtask-conditioned RL (ScRL).}}
ScRL requires the agent to make decisions according to different subtasks. A line of work~\cite{sohn2018hierarchical,sutton2022reward,mehta2005multi,nasiriany2019planning,chane2021goal,ghosh2018learning,zhang2022multi}
focuses on leveraging the prefined or learned abstract subtasks, which lack transparency and have difficulty solving long-horizon problems. There is some work that tends to develop some explainable methods.
From the perspective of the neuro symbol, BPNS \cite{silverlearning} learns neuro-symbolic skills and uses heuristic search A* to guide decision-making.
Pulkit et al.~\cite{verma2022discovering} propose learning user-interpretable capacities from manually collected trajectories and the user's vocabulary. However, how to exploit these subtasks to make long-term planning is not considered, which is very unfavorable for complex tasks.
However, the existing works mainly solve the short-horizon subtask and rarely focus on the recombination of the learned subtask to solve the complicated long-horizon missions.
We propose an explicit subtask representation and scheduling framework to provide long-horizon planning and guidance.
 \vspace{-6pt}

%% file: 3-pre.tex

\section{Problem Statement}


In this section, we will commence by introducing some fundamental concepts, followed by a formal definition of the temporal subtask-conditioned Markov decision process.


\subsection{Definitions}

In this paper, we investigate how to solve a complicated task that contains some simpler parts by representation and planning.
Some fundamental concepts are explained as follows.

\begin{itemize}
    \item \textbf{Subtask}. A subtask is a smaller, more manageable component of a complex task. The completion of the task often requires an orderly combination of multiple subtasks.
    \item \textbf{Subtask representation}. Subtask representation refers to the encoding of the essential information that characterizes a subtask, which is typically learned from the data and used by an agent to facilitate the identification and execution of subtasks during a complex task.
    \item \textbf{Subtask planning}. In reinforcement learning, subtask planning is often formalized as a sequential decision-making problem, where the agent chooses a sequence of subtasks to perform, in order to maximize the cumulative reward over the entire task.
\end{itemize}

\subsection{Problem Setup}

\textbf{Temporal Subtask-conditioned Markov Decision Process (TSc-MDP).}
We consider augmenting a finite-horizon Markov decision process~\cite{howard1960dynamic} $(\mathcal{S}, \mathcal{A}, p, r,{\rho _0}, \gamma) $ with an extra tuple $(\{{\mathcal{T}^i}\}_{i=1}^n,$ $sp,$ ${\rho _s})$ as the temporal subtask-conditioned Markov decision process, where $\mathcal{S}$, and $\mathcal{A}$ denote the state space and the action space, respectively. 
$\{{\mathcal{T}^i}\}_{i=1}^n$ is the subtask set and $n$ is the number of subtasks.
$p(s_{t+1},\mathcal{T}_{t+1}^i \vert s_t,\mathcal{T}_t^i,a_t)$ is the dynamics function that gives the distribution of the next state $s_{t+1}$ and next subtask $\mathcal{T}_{t+1}^i$ at state $s_t$ to execute the subtask $\mathcal{T}_t^i$ and take action $a_t$.
$r:\mathcal{S} \times \mathcal{T} \times \mathcal{A} \to \mathbb{R}$ denotes the reward function. $\rho_0$ and $\rho_s$ denote the initial state distribution and the subtask distribution, respectively. 
$\gamma \in (0,1)$ is the discount factor.
$sp$ is the temporal subtask planner that generates subtask sequences at a coarse time scale. 
Formally, the optimization objective is to find the optimal subtask planner $sp$ and policy $\pi$ to generate the orderly subtask plans and then make a decision for maximizing the expected cumulative return:
\begin{equation}
    J(\pi ,sp ) = {\mathbb{E}_{\scriptstyle{a_t}\sim \pi ( \cdot  \vert {s_t})\hfill\atop
\mathcal{T}_t^i\sim \rm{sp} ( \cdot  \vert \tau )\hfill}}\left[ {\sum\nolimits_t {{\gamma ^t}r({s_t},\mathcal{T}_t^i,{a_t})} } \right], 
\end{equation}
where $\tau$ is the history transition. In our setting, we focus on how to extract the subtask representation and then devise a rational subtask execution plan for the AI agent to make efficient decisions.

%% file: 4-method.tex
\section{Methodology}

\revise{
In this section, we begin by articulating the underlying motivation that guided the development of our approach. Following this, we elaborate on our unique method for learning distinguishable subtask representations, serving as the foundational element of our algorithm. Subsequently, we introduce our innovative construction of a top-$K$ subtask planning tree, specifically designed to customize the agent's subtask execution plans. Finally, we discuss how this planning tree facilitates and enhances policy learning for effective decision-making.
}


\subsection{\revise{The Motivation for the Proposed Framework}}


\revise{
Current reinforcement learning methods often face two major challenges when tackling complex tasks: 
(a) Suboptimal Knowledge Representation: Traditional single-encoder models struggle to capture the unique dynamics of individual subtasks. This leads to poor generalization and inefficiency in learning complex tasks.
(b) Lack of Long-term Planning: Current RL algorithms are often myopic, focusing only on immediate rewards. This short-sightedness often leads to suboptimal policies, as agents can get stuck in local optima without considering long-term consequences.

To address the first challenge of suboptimal knowledge representation, we should focus on:
\begin{itemize}
    \item \textbf{Specialization for Each Subtask}: We propose a multiple-encoder architecture to provide specialized learning and representation for each subtask, thereby tackling the inefficiency caused by one-size-fits-all models.
    \item \textbf{Task-Essential Feature Extraction}: An individual predictor accompanying each encoder is designed to capture the unique, task-relevant features, enabling better discrimination between different subtasks.
\end{itemize}

To address the second challenge of the lack of long-term planning, some factors should be taken into consideration:
\begin{itemize}
    \item \textbf{Enabling Forward-Looking Decisions}: We introduce the concept of a top-$K$ subtask planning tree to guide the agent's policy. This allows for longer-term planning by offering a roadmap of potential future subtasks.
    \item \textbf{Dynamic Scalability}: The top-$K$ subtask planning tree is not fixed in its breadth or depth, offering flexibility in addressing tasks of varying complexity.
    \item \textbf{Global Optima over Local Optima}: By giving the agent the ability to consider a range of future possibilities, the planning tree aids the agent in making decisions that are globally optimal rather than just locally optimal.
\end{itemize}

By addressing these identified challenges, our proposed framework aims to fill the existing gaps in the domain of RL, providing a more effective and comprehensive solution for complex decision-making tasks.
In the subsequent sections, we will give detailed descriptions about three core components of our methodology:
(a) the extraction of distinguishable subtask representations through a multiple-encoder and individual-predictor regime, (b) the generation of a top-$K$ subtask planning tree, and (c) the formulation of a tree-based policy for better decision-making.

}

\subsection{Distinguishable Subtask Representations}
\label{subsec:task-repre}
{\textbf{Data Collection.}} 
\revise{Due to the time-consuming and often ineffective training for complex tasks, directly collecting expert data is intractable. Therefore, we disassemble the complex task into a set of multiple simple tasks, denoted as $\mathcal{T}: \{\mathcal{T}^1, \mathcal{T}^2,..., \mathcal{T}^n\}$. Simple tasks can easily be trained from scratch to obtain optimal policies.
To collect diverse and relevant data and eliminate the effect of the behavior policies}, we separately collect 3000 trajectories for each subtask, of which 1/3 is collected from the converged PPO~\cite{schulman2017ppo}, 1/3 is collected from the intermediate trained checkpoint of PPO, and the last 1/3 is from the random policy. The data statistics are shown in Table~\ref{tab:data}. 
\revise{The collected data format for subtask $\mathcal{T}^i$ can be denoted as $\{(s_t,a_t,r_t,s_{t+1})_{t=1}^L\}^i$, where $L$ is the trajectory length.} 


\begin{table}[h!]
    \centering
    \begin{tabular}{ccc}
       \toprule
       Subtask & \#Total Episodes & \#Total Transitions \\
       \midrule
       PickupLoc & 3000 & 210375 \\
       PutNext & 3000  & 559201 \\
       Goto & 3000  & 218191  \\
       Open & 3000  & 149349   \\
       \bottomrule
    \end{tabular}
   \caption{The data collection statistics}
   \label{tab:data}
\end{table}

\begin{figure}[htbp]
    \centering
      \includegraphics[width=1.0\linewidth]{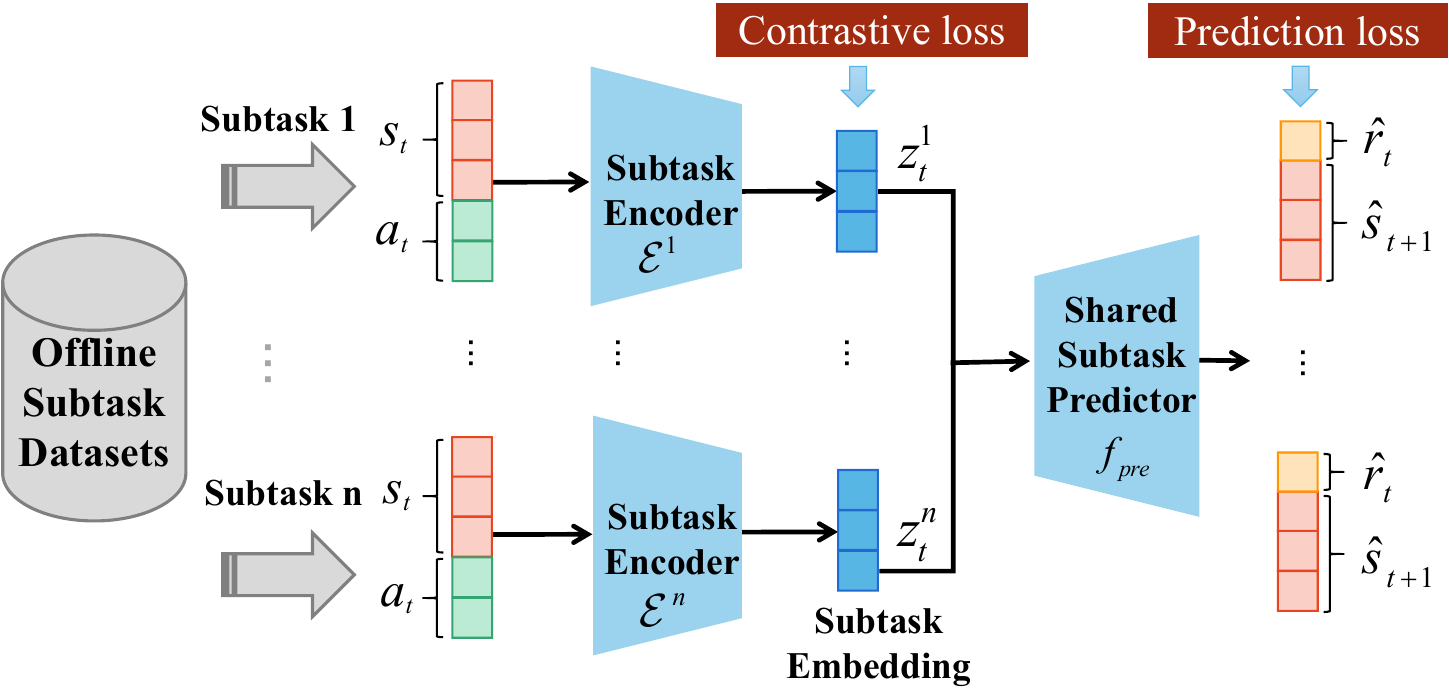}
\caption{The multiple-encoder and individual-predictor learning regime. Each subtask encoder $ \mathcal{E}^i$ encodes the state-action pair $(s_t, a_{t})^i \in \mathcal{T}^i$ and outputs the compact subtask representation $z_t^i$ at each time step. The shared subtask predictor predicts the reward $r_t$ and the next state $s_{t+1}$. The contrastive and prediction losses are designed to train the subtask encoders and the predictor end-to-end.}
	  \label{fig:task-repre}
\end{figure}

{\textbf{Representation Pretraining.}} In reinforcement learning, the environment is modelled by a transition function and a reward function. We use the collected trajectories to extract the essential properties of subtasks and create proper representations based on these functions. To achieve this, we introduce a multiple-encoder and individual-predictor learning regime that can effectively capture the task-essential properties, as shown in Fig.~\ref{fig:task-repre}.

Our framework incorporates two design principles. First, we use multiple encoders to enhance diversity among subtasks, as each subtask has its own unique responsibilities and skills. This approach also ensures that state-action pairs from different subtasks are not mapped to similar abstractions. Second, we introduce an auxiliary prediction task to simulate reward and transition distributions, allowing us to learn accurate and generalizable subtask representations. 
By predicting the reward and next state, our shared subtask predictor implicitly learns the world model, facilitating a better understanding of the essential properties of the subtasks.


Supposed that we have $n$ subtasks, we initialize $n$ subtask encoders parameterized by $\{\theta^i\}_{i=1}^n$, respectively. 
First, we sample a minibatch of $b$ transitions $\{\tau^i_j = (s_j,a_j,r_j,s_{j+1})^i\}_{j=1}^b$ from some subtask $\mathcal{T}^i$, and feed the state-action pairs $\{(s_j,a_j)^i\}_{j=1}^b$ to the subtask encoder $\mathcal{E}^i:\mathcal{S} \times \mathcal{A} \mapsto \mathcal{Z}$. Then the latent variables $z^i \in \mathcal{Z}$ for subtask $\mathcal{T}^i$ are obtained, denoted as $z^i = \mathcal{E}^i((s, a)^i)$. 


For better distinguishment, the contrastive loss is applied. Similar to ~\cite{yuan2022corro,oord2018representation}, we give a definition of anchors, positive, and negative samples.
Using the latent variable $q = {\mathcal{E}^i}(s,a)$ from the subtask $\mathcal{T}^i$ as an anchor, the $k^+ = {\mathcal{E}^i}(s^+,a^+)$ and $k^- = {\mathcal{E}^j}(s^-,a^-)$ respectively constitute positive and negative instances, where $(s^+,a^+)$ and $(s^-,a^-)$ are sampled from the subtask $\mathcal{T}^i$ and other subtasks $\mathcal{T}^j, j \ne i$.
Thus, the contrastive objective~\cite{oord2018representation} is:
\begin{equation}
    {\mathcal{L}_c^i} =  - \log \frac{{\exp ({q} \cdot {k^ + })}}{{\exp ({q} \cdot {k^ + }) + \sum_{k^-} {\exp ({q} \cdot {k^ - })} }}. 
\end{equation}
\revise{
The full set $\mathbb{X}=\{x_0, x_1,...\}$ can be explicitly partitioned as $P(\mathbb{X})=\{\{x^+\}, \{x^-\}\}$, where $x^+$ is from the current subtask, $x^-$ comes from the other $n-1$ subtasks, and $\{x^-\}$=$\mathbb{X} \backslash \{x^+\}$. 
Benefiting from this, the multiple transition encoders should extract the essential variance of transition and reward functions across different tasks and mine the shared and compact features of the same subtasks.
}

Then adopting mini-batch sampling for training, we can optimize the contrastive objective to extract the essential variance of transition functions across different tasks and mine the shared and compact features of the same subtasks.


As for the shared subtask predictor $f_{pre}$, we use the mean squared error (MSE) loss to calculate the mean squared differences between true and prediction as follows.
\begin{equation}
    {\mathcal{L}_p} = {\lambda _r}\|\hat r - r\|_2^2 + {\lambda _s}\|\hat s' - s'\|_2^2   \ ,
\end{equation}
where $\hat r, \hat s' = f_{pre}(z)$ is the prediction of the immediate reward $r$ and the next state $s'$. $\lambda_r$ and $\lambda_s$ are positive scalars that tune the relative weights.

Thus, the proposed framework includes a total loss that combines the contrastive loss, which encourages similar subtask representations for state-action pairs from the same subtask and distinct representations for those from different subtasks, and the dynamics prediction loss, which aims to capture the essential properties of the transition and reward functions, which is summarized as follows.
\begin{equation}
    \min \sum\limits_{subtask \ i} {\sum\limits_{\scriptstyle{\tau^i}\sim{\mathcal{T}^i}\atop
\scriptstyle{\tau^{\{ n\} \backslash i}}\sim{\mathcal{T}^{\{ n\} \backslash i}}} {(\mathcal{L}_c^i + {\mathcal{L}_p})} } , 
\label{eq:pre-train}
\end{equation}
where $\tau=(s,a,r,s')$, and $\{n\}$ are the collected transitions and the total subtask set.

\revise{
In summary, by disassembling complex tasks into simpler, manageable pieces, our approach enables the collection of diverse, various-quality data for each subtask. These data are then channeled through a novel multi-encoder and single predictor paradigm, designed to adeptly grasp the unique dynamics of each subtask, laying a strong foundation for advanced planning and decision-making modules.
The representation pretraining fortifies each encoder's specialization in their respective subtasks, ensuring enhanced, distinct learning and representation. This comprehensive approach guarantees a robust and compact subtask representation for downstream decision-making.
After finishing the first step toward compact representations for subtasks, the subtask tree can be generated for the planning and decision-making process.
}


\subsection{Top-K Subtask Planning Tree Generation}
\label{sec:tree_gen}

\begin{figure*}[!ht]
    \centering
	\subfloat[\small The framework]{
	  \label{fig:task-tree}
	  \resizebox{0.44\textwidth}{!}{
      \includegraphics[height=2.5cm]{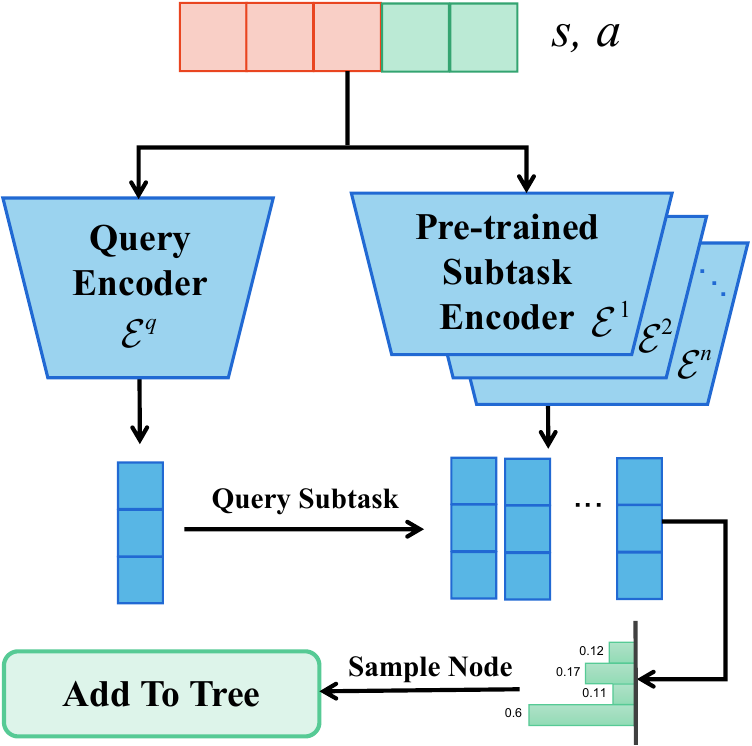}} 
      }
    \subfloat[\small The pseudo code]{
	  \label{fig:vis-base3}
	  \resizebox{0.55\textwidth}{!}{
      \includegraphics[height=2.5cm]{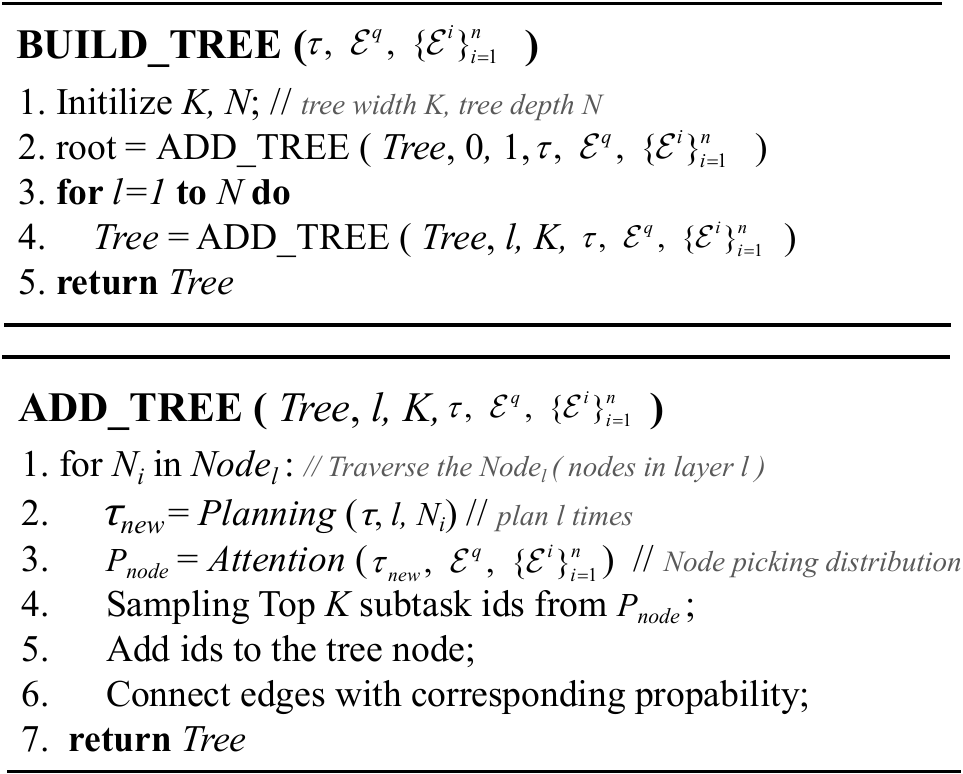}}
      \label{fig:build-tree}
      }
	  \begin{center}
	     \caption{ The generation of the top-$K$ subtask planning tree\protect\footnotemark. 
	  } 
	  \end{center}
\end{figure*}

Effective decision-making in human society often involves considering the current state, predicting the future, and imagining various outcomes. Inspired by this, we propose a method to generate a top-$K$ subtask tree to assist with policy training. To handle long horizons and sparse feedback in complex tasks, we dynamically adjust the subtask tree generation during one episode instead of using a predefined tree. Specifically, we generate a subtask tree in a coarse-to-fine manner with some inference of dynamics, as illustrated in Fig.~\ref{fig:task-tree}. Then we regenerate a new subtask tree once the current tree has been executed to a leaf node. 
Next, we describe the generation process of the node and edge in detail as follows.

\revise{First, we provide the detailed node-edge construction procedure of the subtask tree in a top-down manner. Here, we have a query encoder $\mathcal{E}^q$ and the $n$ pretrained encoder $\{\mathcal{E}^i\}^n_{i=1}$ corresponding to each subtask defined in section~\ref{subsec:task-repre}.}



{\textbf{Node-edge construction.}} 
Here, we have $n$ pretrained encoder $\{\mathcal{E}^i\}^n_{i=1}$ corresponding to each subtask defined in section~\ref{subsec:task-repre}.
We give a query encoder $\mathcal{E}^q$ to retrieve the subtask, which serves as the node in the tree.
Given the last state-action pair $(s_{t-1}, a_{t-1})$, we can compute the query vector $v^q$ and the subtask embeddings $\{ {v^i}\} _{i = 1}^n$ as follows.
\begin{equation}
\label{eq:foot_begin}
    {v^q} = {{{{\mathcal{E}}}}^q}({s_{t - 1}},{a_{t - 1}}) , \ \ \ \ \ 
    \{ {v^i}\} _{i = 1}^n = \{ {{\mathcal{E}}^i}({s_{t - 1}},{a_{t - 1}})\} _{i = 1}^n.
\end{equation}


\footnotetext{Note: \textit{Attention} function is defined by Equation (\ref{eq:foot_begin}-\ref{eq:att3}) and \textit{Planning} function is defined by the $m$-step predictor $p_m$.}

Then we utilize the attention mechanism~\cite{bahdanau2015attention} to compute the correlation score between $v^q$ and $\{ {v^i}\} _{i = 1}^n$. Then the normalized coefficient $\alpha_{q, i}$ is defined:
\begin{equation}
\label{eq:att1}
    {\alpha _{q,i}} = \frac{{\exp ({\beta _{q,i}})}}{{\sum\nolimits_j {\exp ({\beta _{q,j}})} }}, \ where \ {\beta _{q,i}} = {v^q}{W^T_\phi}W_{\varphi}{v^i} , 
\end{equation}
where $W_\phi, W_\varphi$ are learnable parameters and $\beta _{q,i}$ is the attention score. Here, we can model the correlation between query vector $v^q$ and an arbitrary number of subtasks, which allows us to easily increase the number of subtasks in our tree generation process. 
Then the categorical distribution can be modelled using the weights from which the variable $C$ can be sampled:
\begin{equation}
    \label{eq:att2}
    C \sim {\mathcal{P}_\beta } = Categorical\left( {\frac{{\exp ({\beta _{q,i}})}}{{\sum\nolimits_{j \in \mathcal{N} } {\exp ({\beta _{q,j}})} }}, i \in \mathcal{N}=\{ 1,...,n\} } \right).
\end{equation}

Then, given the hyperparameter $K<n$, we can sample the top-$K$ subtask ids without replacement from the categorical distribution which means sampling the first element, then renormalizing the remaining probabilities to sample the next element, etcetera.
Let $C_1^*,...,C_K^*$ be an (ordered) sample without replacement from the Categorical $\mathcal{P}_\beta$, then the joint probability can be formulated as follows.
\begin{equation}
\label{eq:att3}
    P(C_1^* = c_1^*,...,C_K^* = c_K^*) = \prod\limits_{i = 1}^K {\frac{{\exp ({\beta _{q,c_i^*}})}}{{\sum\nolimits_{\iota  \in \mathcal{N}_i^*} {\exp ({\beta _{q,\iota }})} }}} , 
\end{equation}
where $c_1^*,...,c_K^* = argtop $-$ K({\mathcal{P}_\beta })$\protect\footnotemark  is the sampled top-$K$ realization of variables $C_1^*,...,C_K^*$, and $\mathcal{N}_i^* = \mathcal{N}\backslash \{ c_1^*,...,c_{i - 1}^*\} $ is the domain (without replacement) for the $i$-th sampled element.
\footnotetext{$argtop $-$ K(\mathcal{P}_\beta)$ means the corresponding subtask ids of top k probabilities from $\mathcal{P}_\beta$. }

Then, we use the sampled top-$K$ subtask ids as the $K$ nodes at a certain layer in the subtask tree (please note that we set $k=1$ while generating the root node). Moreover, given the hyperparameter $N$, we can obtain the tree of depth $N$ by repeating the above process based on the planning transition (described in the next paragraph), and the original correlation weight ${\alpha _{q, i}}$ is set as connections (edges) between layers.
For a clear understanding, we summarize the pseudo-code shown in Fig.~\ref{fig:build-tree}. 

{\textbf{Planning for future: $m$-step predictor}.}
While expanding the subtask tree, we should make planning for the $m$-step prediction of the future. 
We design an $m$-step predictor denoted as $p_m(\hat s_{t+m}  \vert  \hat s_{t}, \hat a_t)$ that is used to derive the next $m^{th}$ step's state. The action $\hat a_{t+m}$ is output by the policy network $\pi$ (defined in Equation~(\ref{eq:pi})) using the current subtask representation.
The $N$-times rollout starting with $ s_{t-1}$ is as follows:
\begin{equation}
    \left\{ {({s_{t - 1}},{a_{t - 1}}),({{\hat s}_{t + m - 1}},{{\hat a}_{t + m - 1}}),...,({{\hat s}_{t + N \cdot m - 1}},{{\hat a}_{t + N \cdot m - 1}})} \right\}.
\end{equation}
Then we can use this imagined rollout to expand the tree to a greater depth. That is, as each layer of depth increases, the planning will perform $m$-step imagination.


{\textbf{Optimization for tree generation}.}
The optimization includes two parts: an $m$-step predictor $p_m$ and an attention module. The former $p_m$ can be trained by minimizing the mean squared error function $\mathcal{L}_m$ defined as follows. By sampling some state-action pairs from replay buffer $\mathcal{D}$, we have:
\begin{equation}
\mathcal{L}_m = {\mathbb{E}_{\scriptstyle({s_t},{a_t},{s_{t + m }})\sim \mathcal{D}\hfill\atop
\scriptstyle{{\hat s}_{t + m }} = {p_m}({s_t},{a_t})\hfill}}\left[ { \|{{\hat s}_{t + m }},{s_{t + m }} \|_2^2} \right]. 
\label{eq:tree1}
\end{equation}

The attention module formulated by Equations~(\ref{eq:att1}), (\ref{eq:att2}), and (\ref{eq:att3}), and parameterized by $\omega=\{\phi, \varphi\}$ can be optimized in a Monte-Carlo policy gradient manner~\cite{williams1992simple}. 
First, we define the intra-subtask cumulative reward $R_t=\sum\nolimits_{j = 0}^{\tilde T - 1} {{\gamma ^j}{r_{t + j}}}$ during the interval $\tilde T$ of the subtask execution process as follows. By decomposing the cumulative rewards into a Bellman Expectation equation in a recursive manner, we acquire:
\begin{equation}
G({s_t},{\mathcal{T}_t}) = \mathbb{E}\left[ {R_{t+1}  + {\gamma }G({s_{t + \tilde T}},{\mathcal{T}_{t + \tilde T}}) \vert {s_t},{\mathcal{T}_t}} \right], 
\end{equation}
where ${s_t}$ and ${\mathcal{T}_t}$ are the current state and the executed subtask, respectively. $\gamma$ is the discounted factor, and ${\mathcal{T}_{t+ \tilde T}}$ is the next subtask. 

Then, by applying the mini-batch technique to the off-policy training, the gradient updating can be approximately estimated as:
\begin{equation}
    \omega \leftarrow \omega + \alpha \nabla \log {\mathcal{P}_\beta }({\mathcal{T}_t} \vert {s_{t-1}, a_{t-1} }) \cdot {G(s_t, \mathcal{T}_t)}, 
\label{eq:tree2}
\end{equation}
where $t$ denotes the timestep when a subtask execution begins and $\alpha$ is the learning rate.

\revise{

In summary, our top-$K$ subtask planning tree module revolutionizes decision-making by dynamically creating subtask trees to assist in policy training. Inspired by human-like foresight, we adjust the subtask tree during episodes, rather than relying on predefined structures, effectively handling long horizons and sparse feedback in complex, dynamic tasks. 
}

\subsection{Tree-auxiliary Policy}
\label{sec:d-ucb}

To obtain a tree-auxiliary policy, we start by generating a plan from the top-$K$ subtask tree. This plan serves as a guideline for decision-making, enabling agents to make rational actions based on a foresight judgment of the situation.


{\textbf{Subtask execution plan generation}.}
We propose a discounted upper confidence bound (d-UCB) method for pathfinding, where we record the selection count $c_{\mathcal{T}^i}$ and the average cumulative rewards $r_{\mathcal{T}^i}$ for each subtask $\mathcal{T}^i$ while terminating.
The UCB value for the $\mathcal{T}^i$ in the planning tree is computed as:
\begin{equation}
    v_{\mathcal{T}^i} = r_{\mathcal{T}^i} + 0.5 \cdot \sqrt {\frac{2\ln c_{{\mathcal{T}^i}}^{pa}}{c_{\mathcal{T}^i}}}, 
\end{equation}
where $c_{{\mathcal{T}^i}}^{pa}$ is the selection count of the parent node of $\mathcal{T}_i$.


Then, we define the discounted path length using d-UCB and then use the greedy policy to select the path with the maximum discounted length. More details behind the d-UCB design can be found in Appendix~\ref{app:d-uct}.

\begin{definition}
Sample a path from the root to leaf in the tree, the length of the discounted path can be defined as follows.
\begin{equation}
    {{\hat d}}(root \to leaf) = \sum\nolimits_l {{\kappa ^l}{e^{l,l + 1}} \cdot v_{\mathcal{T}^i}}, 
\end{equation} 
where $e^{l,l+1}$ is a sampled edge between the $l^{th}$ layer and $(l+1)^{th}$ layer, the value of the edge is the probability of subtask transition, $\kappa=0.8$ is the discounted factor, and $v_{\mathcal{T}^i}$ is the UCB value of the sampled node in the $(l+1)^{th}$ layer.
\end{definition}

{\textbf{Subtask termination condition}.} 
Subtask execution should be terminated when the current situation changes, and we provide a measurement about when to terminate the current subtask. 
We first calculate the cosine similarity of the current query vector $\mathcal{E}^q(s_t,a_t)$ and the corresponding subtask embedding $\mathcal{E}^{\mathcal{T}_{id}}(s_t,a_t)$ and the similarity function is defined as follows.
\begin{equation}
Sim({{{{\mathcal E}}}^q}({s_t},{a_t}),{{{{\mathcal E}}}^{{{{{\mathcal T}}}_{id}}}}({s_t},{a_t})) = \frac{{\left( {{{({{{{\mathcal E}}}^q}({s_t},{a_t}))}^T}{{{{\mathcal E}}}^{{{{{\mathcal T}}}_{id}}}}({s_t},{a_t})} \right)}}{{\left( {||{{{{\mathcal E}}}^q}({s_t},{a_t})|| \cdot ||{{{{\mathcal E}}}^{{{{{\mathcal T}}}_{id}}}}({s_t},{a_t})||} \right)}}. 
\end{equation}

For every timestep, the agent will compare the similarity score of the current timestep and the last timestep. 
Avoiding the drastic changes in subtasks, we choose $\Delta = 0.5$ as the division.
A similarity score of less than $0.5$ indicates a significant change in the environment, and the current subtask should terminate. 


\begin{algorithm}[ht!]
\caption{Overall Algorithm Procedure}
\label{algo:nn}
\begin{algorithmic}[1]
\Ensure subtask representation policy $\rho=\{ \{\mathcal{E}^i\}_{i=1}^n, f_{pre}\}$, tree building policy $\mathcal{E}^q$, and tree-axuliary policy $\pi$; 
\State // \textit{Subtask Representation Pretraining} 
\State Collect diverse subtask datasets $\{\mathcal{T}^1, ..., \mathcal{T}^n\}$; 
\For{\textit{each iteration}}
    \State Sample a subtask $\mathcal{T}^i$ and transitions $\{\tau^i_t\}_{t=1}^b$, $\{\tau^{i,+}_t\}_{t=1}^b$;
    \State $\{z^i_t\}_{t=1}^l = \{\mathcal{E}^i(s_t^i, a_t^i)\}_{t=1}^l$ ; 
    \State $\{z^{i,+}_t\}_{t=1}^l = \{\mathcal{E}^i(s_t^{i,+}, a_t^{i,+})\}_{t=1}^l$ ; 
    \State Sample other subtasks $\mathcal{T} \backslash \mathcal{T}^i$ and transitions $\{\tau^{i,-}_t\}_{t=1}^b$; 
    \State $\{z^{i,-}_t\}_{t=1}^l = \{\mathcal{E}^{-}(s_t^{i,-}, a_t^{i,-})\}_{t=1}^l$ ; 
    \State Predict reward $r$ and next state $s'$;
    \State Update $\rho$ to minimize Eq.~(\ref{eq:pre-train});
\EndFor
\State // \textit{Tree Generation and Tree-auxiliary Policy} 
\For{\textit{each episode}}
\State Initial state-action pair $s\_a_{cur} = (s_0,a_0)$; 
\State BUILD\_TREE($s\_a_{cur}, \mathcal{E}^q, \{\mathcal{E}^i\}_{i=1}^n$) ; // \textit{Fig.~\ref{fig:build-tree}} 
\State Use d-UCB in Sec.~\ref{sec:d-ucb} to find a plan $\zeta$;
\For{\textit{each timestep} $t$}
\If{ $\zeta$ is None}
\State Regenerate the tree and the plan; 
\EndIf
\While{$\zeta$ is not None}
\State $\mathcal{T}_{id} = \zeta[0]$;
\State // \textit{execute current subtask}
\State Select action $a_t = \pi (\cdot  \vert s_t, \mathcal{E}^{\mathcal{T}_{id}}(s\_a_{cur}))$;
\State Receive reward $r$ and next state $s'$ ; 
\If{ $Sim({{{{\mathcal E}}}^q}({s_t},{a_t}),{{{{\mathcal E}}}^{{{{{\mathcal T}}}_{id}}}}({s_t},{a_t}))<0.5$ 
}
\State remove $\mathcal{T}_{id}$ from $\zeta$; // \textit{execute next subtask} 
\EndIf
\EndWhile
\EndFor
\State Update $\pi$ with Eq.~(\ref{eq:pi}); 
\State Update $\mathcal{E}^q$ with Eq.~(\ref{eq:tree1}) and Eq.~(\ref{eq:tree2});
\EndFor
\end{algorithmic}
\end{algorithm}



{\textbf{Subtask-guided decision making}.}
In our implementation, the policy will consider the current state and the subtask embedding and give a comprehensive decision, formulated as follows.
\begin{equation}
    a_t \sim \pi(\cdot  \vert  s_t, \mathcal{E}^{\mathcal{T}_{id}}(s_{t-1},a_{t-1})). 
    \label{eq:pi}
\end{equation}

As for the policy optimization, we adopt the PPO-style~\cite{schulman2017ppo} training regime. We summarized the objective function as follows.
\begin{equation}
    {\mathcal{L}_\pi } = \left[ {\min \left( {{i_t} \cdot {A_t},clip({i_t},1 - \epsilon ,1 + \epsilon ){A_t}} \right)} \right], 
    \label{eq:pi-loss}
\end{equation}
where $i_t = \frac{{\pi ({a_t} \vert {s_t},{{{{\mathcal E}}}^{{{{{\mathcal T}}}_{id}}}}(s_{t-1},a_{t-1})  )}}{{{\pi _{old}}({a_t} \vert {s_t}, {{{{\mathcal E}}}^{{{{{\mathcal T}}}_{id}}}}(s_{t-1},a_{t-1})  )}}$ 
\revise{represents the importance ratio between the current policy $\pi$ and previous policy $\pi_{old}$, which corresponds to the policy employed in the last training step.}
$A_t$ is an estimator of the advantage function at timestep $t$.
\revise{The parameter $\epsilon = 0.2$ serves as the clipping threshold in the PPO algorithm. The {clip($\cdot$)} function is utilized to constrain its value of $i_t$ within the interval $[1 - \epsilon, 1 + \epsilon]$. This function serves a critical role in mitigating excessively aggressive updates to the policy, thereby contributing to the overall stability and efficiency of the algorithm.}

The overall optimization flow including subtask representation pretraining, top-$K$ subtask generation, and tree-auxiliary policy is summarized in Algorithm~\ref{algo:nn}.

\revise{

In summary, the tree-auxiliary policy leverages the top-$K$ subtask planning tree generated through the dynamic decision-making process. This plan serves as a valuable guideline for agents, allowing them to make informed decisions based on forward-looking assessments of the current scenario.
Incorporating subtask representations into the policy conditions broadens the agent's perspective, enabling effective policy adjustments. This tree-auxiliary policy component complements our framework, contributing to its adaptability and robust decision-making capabilities.

}

\revise{

{\subsection{Overall Analysis of Proposed Modules}} 
In this subsection, we will elaborate on how our proposed Top-$K$ Subtask Planning Tree can tackle long-horizon missions and how the combination of subtask representations and the planning tree contributes to solving this particular challenge.

\textbf{Question 1: Why do top-$K$ subtask planning tree resolve long-horizon missions?}  

In long-horizon tasks where agents must plan and execute a series of actions, traditional reinforcement learning methods face a hard challenge as they typically struggle with sparse feedback and signals spanning extensive time steps. Our top-$K$ subtask planning tree addresses this by fragmenting larger tasks into smaller, more manageable subtasks. Each subtask operates within a shorter timespan, receiving more frequent and explicit feedback, significantly simplifying the learning and decision-making process.

\textbf{Question 2: What is the role of subtask representations? }

Subtask representations enable our model to understand and handle tasks in a structured manner. By encoding each subtask, the model can more easily discern relationships and dependencies among tasks, assisting in planning and executing actions more effectively. This approach also offers a pathway for more efficient reuse of previously learned knowledge and skills, accelerating the learning process.

\textbf{Question 3: What is the contribution of the top-$K$ subtask planning tree?}

The planning tree further augments the capabilities of our approach. By utilizing a planning tree, our model can make both long-term and short-term decisions within a unified framework. At each node, the model dynamically generates and adjusts the subtask tree based on the current state and contextual information, achieving flexible and efficient decision-making. The planning tree also affords the model a means to understand and explain its decisions and actions, enhancing its interpretability and transparency.

\textbf{Question 4: Why do we need the combination of these modules?} 

In summary, our top-$K$ subtask planning tree, by integrating subtask representations and a planning tree, effectively addresses the challenges posed by long-horizon tasks. Subtask representations facilitate the model in handling complex tasks with ease, while the planning tree provides a unified framework for efficient and flexible decision-making. The combination of these elements not only empowers our model to successfully accomplish long-horizon tasks but also maintains its explainability and transparency, enhancing its reliability and practicality in real-world applications.

}

%% file: 5-experiments.tex
\section{Experiments}

\revise{In this section, we seek to empirically evaluate our method to verify the effectiveness of our solution. We design the experiments to 1) verify the effectiveness of the multiple-encoder and individual-predictor learning regime; 2) test the performance of our top-$k$ subtask tree on the most complicated scenarios in the 2D navigation task BabyAI~\cite{babyai_iclr19}; 3) conduct some sensitivity analyses in our settings; 4) give some intuitive statistics about our model.}

\subsection{Experimental Setting}

We choose BabyAI as our experimental platform, which incorporates some \textbf{basic} subtasks and some composite \textbf{complex} tasks. 
For each scenario, there are different small subtasks, such as opening doors of different colors. The agent should focus on the skill of opening doors, rather than the color itself.
Moreover, the BabyAI platform provides by default a $7\times7\times3$ symbolic observation $o_t$ (a partial and local egocentric view of the state of the environment) and a variable length instruction $c$ as inputs at each time step $t$.




\begin{figure*}[!h]
    \centering
	\subfloat[\small Open]{
	  \label{fig:vis-base2}
	  \resizebox{.2\textwidth}{!}{
      \includegraphics[height=2.5cm]{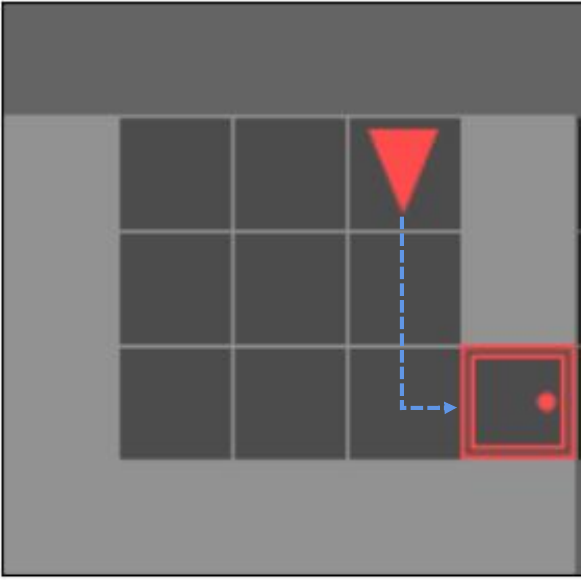}} 
      }
    \subfloat[\small Go to]{
	  \label{fig:vis-base3}
	  \resizebox{.2\textwidth}{!}{
      \includegraphics[height=2.5cm]{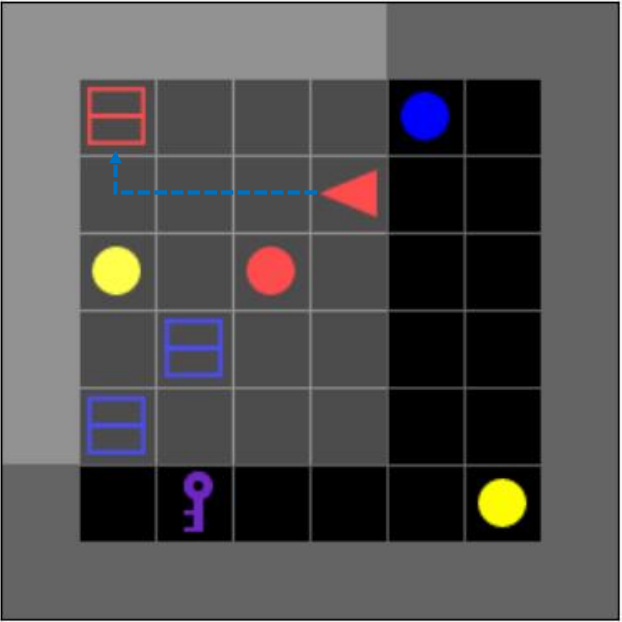}}
      }
    \subfloat[\small Put next]{
	  \label{fig:vis-base4}
	  \resizebox{0.213\textwidth}{!}{
      \includegraphics[height=2.5cm]{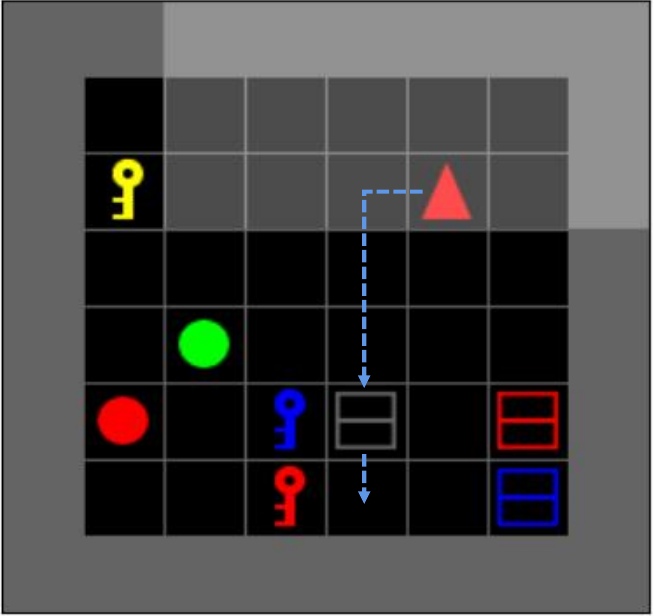}}
      }
      \subfloat[\small Pick up]{
	  \label{fig:vis-base4}
	  \resizebox{0.2\textwidth}{!}{
      \includegraphics[height=2.5cm]{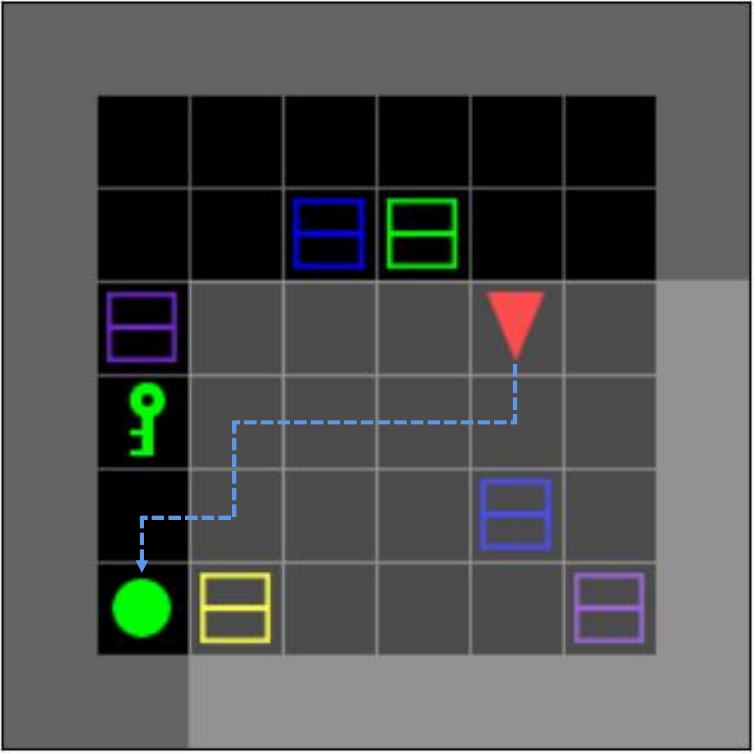}}
      }
	  \begin{center}
	     \caption{The schematics of four basic subtasks. The dashed lines denote the path to reach the goal. The agent represented by the red triangle is partially observable to the grid and the light-grey shaded area represents its field of view. The missions are described as :  (a). Open the red door; (b). Go to the red box; (c). Put the grey box next to the red key. (d). Pick up the green ball. 
	  } 
	  \label{fig:env-base}
	  \end{center}
\vspace{-10pt}
\end{figure*}

{\textbf{Basic Subtask}.}
BabyAI contains four basic subtasks that appear in more complicated scenarios, including: (a) \textit{\textbf{Open}}: Open the door with a certain color and a certain position, e.g. "open the green door on your left". The color set and position set are defined as : $COLOR = \{ red, \ green, $ $blue, \   purple, $ $ yellow, \ grey\}$ and $POS = \{on \ your \ left, \ on \ your \ right,$ $ \ in \ front \ of \ you,  \ behind $  $\  you\}$. (b) \textit{\textbf{Goto}}:   Go to an object that the subtask instructs, e.g. "go to red ball" means going to any of the red balls in the maze. The object set is defined as: $OBJ = \{door, \ box, \ ball, \ key\}$. (c) \textit{\textbf{PutNext}}: Put objects next to other objects, e.g. "put the blue key next to the green ball". (d) \textit{\textbf{PickupLoc}}: Pick up some objects, e.g.``pick up a box''.
The detailed subtask illustrations are shown in Fig.~\ref{fig:env-base}.


\begin{figure}[h!]
    \centering
      \includegraphics[width=0.4\linewidth]{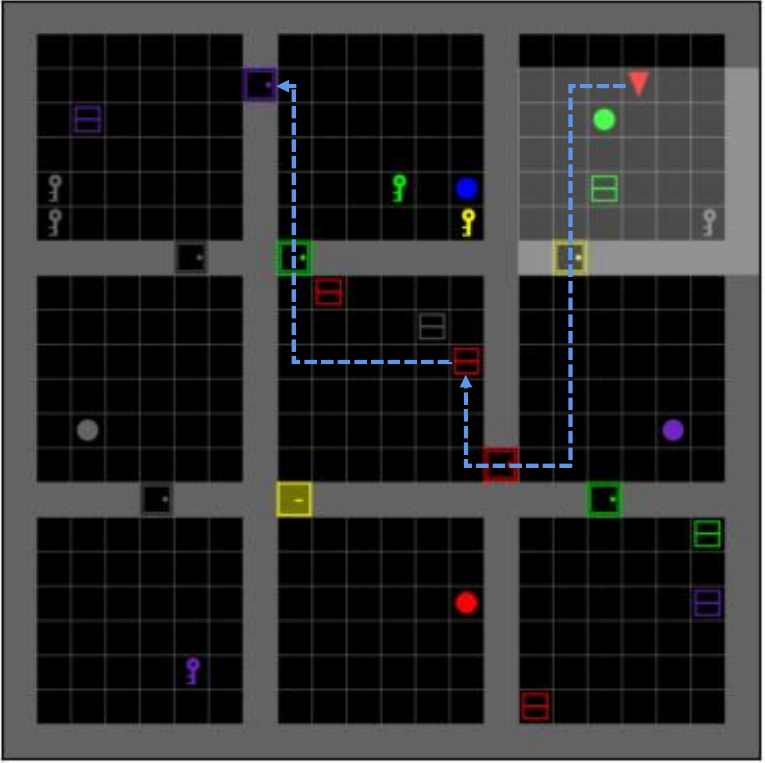}
\caption{\revise{Illustration of one of the BossLevel scenarios. The mission is: ``Pick up a red box and go to the purple door''. }}
	  \label{fig:bosslevel_}
\end{figure}

{\textbf{Complex Task}.}
The BabyAI platform includes some tasks that require three or more subtask compositions to solve, which are called complicated tasks. 
What we want to do is to solve these tasks through our tree-based policy to reach higher performance and faster convergence.
Take the "BossLevel" scenario as an example, the agent should follow one of the instructions "pick up the grey box behind you, then go to the grey key and open a door", which contains three subtasks, \revise{as shown in Fig.~\ref{fig:bosslevel_}.} More detailed descriptions are shown in Appendix~\ref{app:more-complcated}.

{\textbf{Baselines}.}
In this paper, we select the state-of-the-art (SOTA) algorithms PPO~\cite{schulman2017ppo} and SAC~\cite{haarnoja2018sac}, the tree-based planning algorithm SGT~\cite{jurgenson2020sub}, the context-based RL method CORRO~\cite{yuan2022corro} and 
 DGRL~\cite{islam2022discrete}, and LM-based planner LEAP~\cite{chen2023planning} as baselines to verify the effectiveness of our proposed method.
\begin{itemize}
    \item PPO is a popular deep reinforcement learning algorithm for solving discrete and continuous control problems, which is an on-policy algorithm.
    \item SAC is an off-policy actor-critic deep RL algorithm based on the maximum entropy reinforcement learning framework.
    \item SGT adopts the divide-and-conquer style that recursively predicts intermediate subgoals between the start state and goal. It can be interpreted as a binary tree that is most similar to our settings.
    \item CORRO is based on the framework of context-based meta-RL, which adopts a transition encoder to extract the latent context as a policy condition.
    \item \revise{DGRL learns a latent representation for both observations and goals and converts it into discrete latent based on a vector quantized variational autoencoder quantization bottleneck for downstream RL tasks in the context of goal-conditioned hierarchical reinforcement learning.} 
    \item \revise{LEAP utilizes a masked language model to iteratively plan for the decision-making process, leading to improved RL performance across different tasks.}
\end{itemize}

\revise{To ensure the impartiality of our experiments, we pretrain the transition encoder for CORRO using the same subtask transitions that we collected for our own setting. Additionally, the dataset used for LEAP is identical to ours. We have included a table outlining key hyperparameter settings for the baseline models in Table~\ref{tab:para_baseline}, and for more detailed configurations, we refer readers to the original papers.}


\begin{table}[ht!]
    \centering
    \begin{tabular}{ccc}
           \toprule
           \revise{Algorithm} & \revise{Description} &  \revise{Value}  \\
           \midrule
           ~  & \revise{clipping parameter} & \revise{$0.1$}  \\
           ~  & \revise{VF coeff. $c_1$} & \revise{$1$}  \\
           \revise{PPO}  & \revise{ Entropy coeff. $c_2$} & \revise{$0.01$}  \\
           ~  & \revise{ GAE parameter $\lambda$} & \revise{$0.95$}  \\
           ~  & \revise{learning rate} & \revise{$10^{-4}$}  \\
           \midrule
           \revise{SAC } &  \revise{target smoothing coefficient $\tau$ } & \revise{$0.005$ }  \\
           ~  &  \revise{learning rate} & \revise{$3*10^{-4}$ }  \\
           \midrule
           \revise{SGT}  &  \revise{depth $D$ } & \revise{$3$ }  \\
           ~  &  \revise{learning rate} & \revise{$0.005$ }  \\
           \midrule
            ~  &  \revise{negative pairs number} &   \revise{$64$}  \\
           \revise{ CORRO}  &  \revise{task embedding size}  &   \revise{$5$} \\
           ~  &  \revise{learning rate} & \revise{$3*10^{-4}$ }  \\
           \midrule
           ~  &  \revise{ codebook size  $L$ } & \revise{$10$ }  \\
           \revise{DGRL} &  \revise{the number of factors per vector $G$ } & \revise{$16$ }  \\
            ~  &  \revise{task embedding size}  &   \revise{$5$} \\
           ~  &  \revise{learning rate} & \revise{$3*10^{-4}$ }  \\
           \midrule
            ~  &  \revise{embedding size}  &   \revise{$128$} \\
            ~  &  \revise{Instruction Encoder channels}         &    \revise{$128$} \\
            \revise{LEAP}  &  \revise{State Encoder channels}        &     \revise{$[256, 256, 128]]$ }\\
            ~  &  \revise{Image Encoder channels} &    \revise{$[ 128, 128]$ }\\
            ~  &  \revise{learning rate} &    \revise{$6*10^{-4}$ } \\
           \bottomrule
    \end{tabular}
    \caption{\revise{The key hyperparameters of baselines. }}
    \label{tab:para_baseline}
\end{table}



{\textbf{Parameter Settings.}}
Additionally, we provide the detailed experimental hyperparameter settings \revise{shown in Table~\ref{tab:para1_1} and Table~\ref{tab:para2_1}}, and the description of all the neural networks used in our framework is shown in Appendix~\ref{app:para-nn}.

\begin{table}[ht!]
    \centering
    \begin{tabular}{cc}
           \toprule
           \revise{Description} &  \revise{Value}  \\
           \midrule
           \revise{learning rate} & \revise{$3*10^{-4}$}  \\
           \revise{contrastive batch size} & \revise{$64$ }  \\
            \revise{negative pairs number} &   \revise{$16$}  \\
            \revise{task embedding size}  &   \revise{$5$} \\
            \revise{$\lambda_r$}         &    \revise{$0.9$} \\
            \revise{$\lambda_s$}        &     \revise{$1.0$ }\\
            \revise{encoder hidden size} &    \revise{$[256, 256, 128, 128, 32]$ }\\
            \revise{decoder hidden size} &    \revise{$[64, 128]$ } \\
           \bottomrule
    \end{tabular}
    \caption{\revise{The hyperparameters of the module for the multiple-encoder and individual-predictor regime. }}
    \label{tab:para1_1}
\end{table}

\begin{table}[ht!]
    \centering
       \begin{tabular}{cc}
           \toprule
           \revise{Description} &  \revise{Value} \\
           \midrule
           \revise{optimizer  }     &   \revise{ Adam }  \\
            \revise{$\alpha$  }      &  \revise{  $10^{-4}$ } \\
            \revise{$\beta_1$ }      &  \revise{  $0.9$ } \\
            \revise{$\beta_2$}       &  \revise{  $0.999$}  \\
            \revise{$\varepsilon$}   &  \revise{  $10^{-5}$}  \\
            \revise{seed}             &  \revise{ $[0,10)$  }\\
            \revise{number of process} & \revise{ $64$ } \\
            \revise{total frames}   &    \revise{ $10M$ } \\
            \revise{learning rate}            &    \revise{ $10^{-4}$ } \\
            \revise{batch size }    &    \revise{ $1280$ } \\
            \revise{eval interval}  &    \revise{ $1$ } \\
            \revise{eval episodes}  &    \revise{ $500$ }  \\
            \revise{image embedding} &   \revise{ $128$ } \\
            \revise{memory of LSTM}   &  \revise{ $128$ } \\
            \revise{query hidden size} &  \revise{$[32, 16, 8]$ } \\
            \revise{number of head  }        &  \revise{ $1$ } \\
            \revise{discounted factor of UCB} &\revise{ $0.9$ }\\
            \revise{$m$-step predictor} & \revise{$5$} \\
           \bottomrule
    \end{tabular}
    \caption{\revise{The hyperparameters of the module for the top-K subtask planning tree. }}
    \label{tab:para2_1}
\end{table}

\begin{figure}[h!]
\centering
	\subfloat[\small 2 subtasks]{\includegraphics[width = 0.26\textwidth]{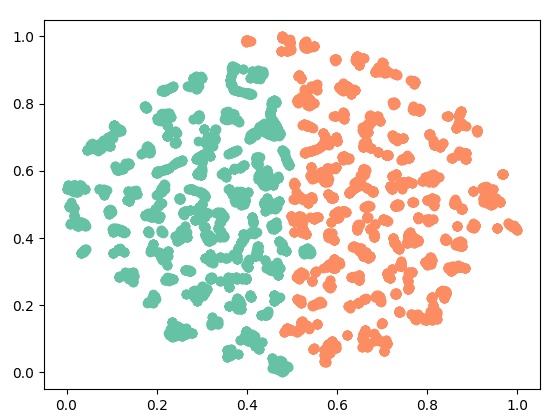}}
	\subfloat[\small 3 subtasks]{\includegraphics[width = 0.26\textwidth]{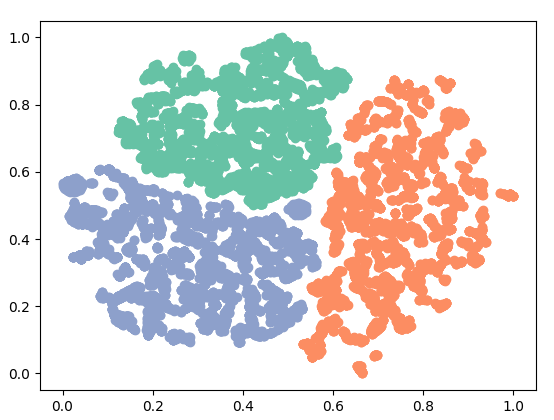}}
	\subfloat[\small 4 subtasks]{\includegraphics[width = 0.26\textwidth]{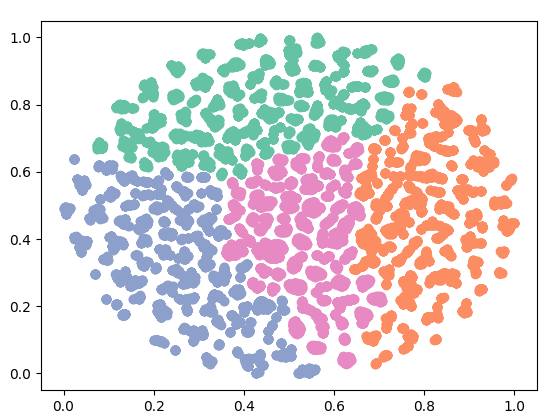}} 
	\hfill
 
	\subfloat[\small 2 subtasks]{\includegraphics[width = 0.26\textwidth]{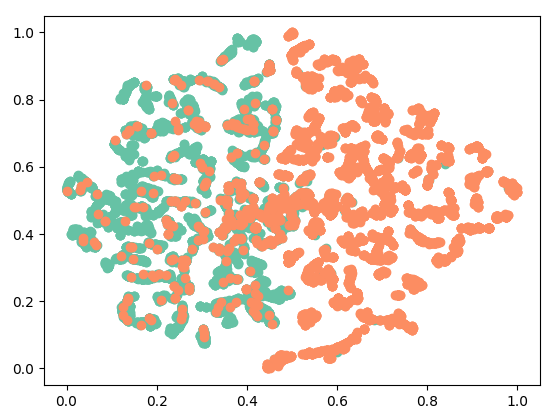}}
	\subfloat[\small 3 subtasks]{\includegraphics[width = 0.26\textwidth]{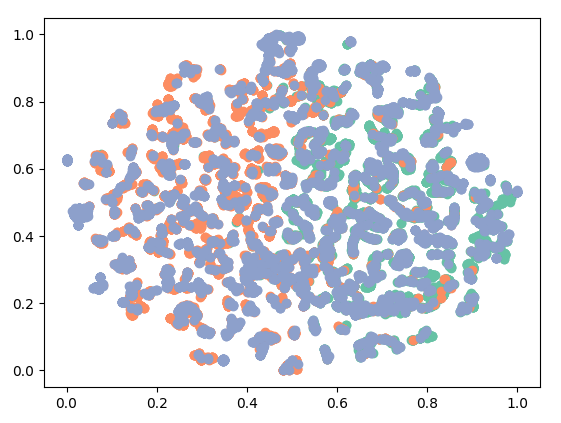}}
	\subfloat[\small 4 subtasks]{\includegraphics[width = 0.26\textwidth]{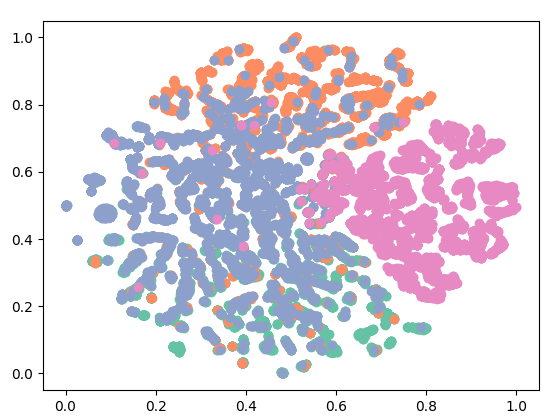}} 
	
\caption{The t-SNE visualizations of subtask embeddings. (a-c): The results are generated by our multiple encoders and one predictor learning regime on 2, 3, and 4 subtasks, respectively. (d-f): The visualization for one shared encoder and one decoder framework.}
\label{fig:exp-1}
\end{figure}

\subsection{Results on Subtask Representation}

We present visualization results in Fig.~\ref{fig:exp-1} to validate the effectiveness of our subtask representations described in Section~\ref{subsec:task-repre}. 
Specifically, we compare our approach with an ablated version where a shared encoder is used instead of multiple encoders for each subtask. We train both models on 2, 3, and 4 subtasks, respectively, and plot the t-SNE results of subtask embeddings in Fig.~\ref{fig:exp-1}. The results in Fig.~\ref{fig:exp-1}(a-c) demonstrate that our approach can clearly differentiate separate clusters of embeddings when using 2, 3, and 4 encoders for each subtask. However, Fig.~\ref{fig:exp-1}(d-f) shows that using only one shared encoder cannot effectively learn the boundaries and distinguish subtasks. We conjecture that training subtasks with mixed representations may interfere with the stabilization of the representation module.

Moreover, due to the powerful capacity of contrastive learning and model-based dynamic prediction, our model can exhibit strong representational discrimination and contain enough dynamic knowledge, which will be verified empirically as follows.

\begin{figure*}[ht!]
\centering
	\subfloat[\small GoToSeq]{\includegraphics[width = 0.314\textwidth]{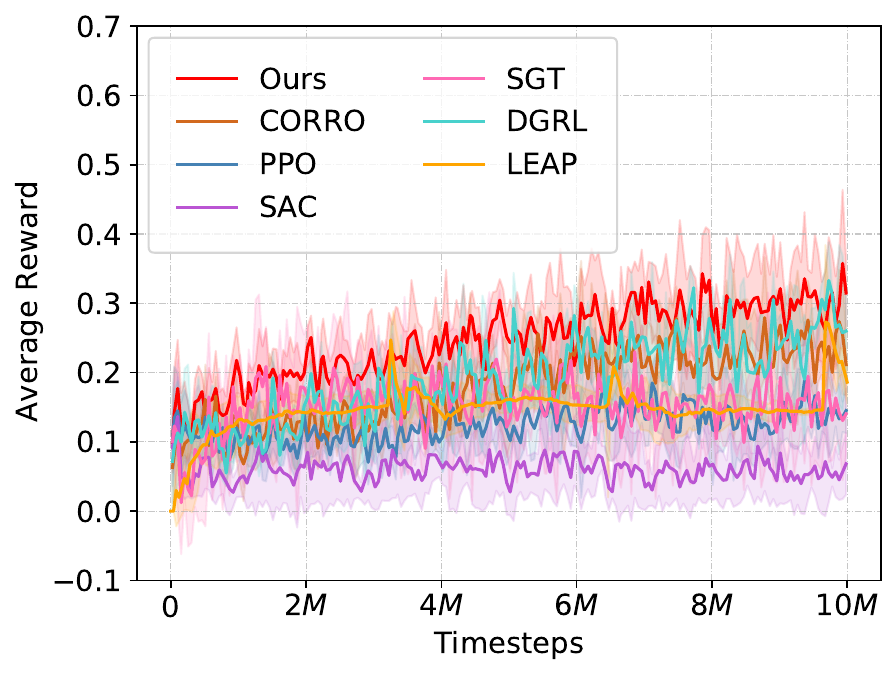}}
	\subfloat[\small GoToSeqS5R2]{\includegraphics[width = 0.31\textwidth]{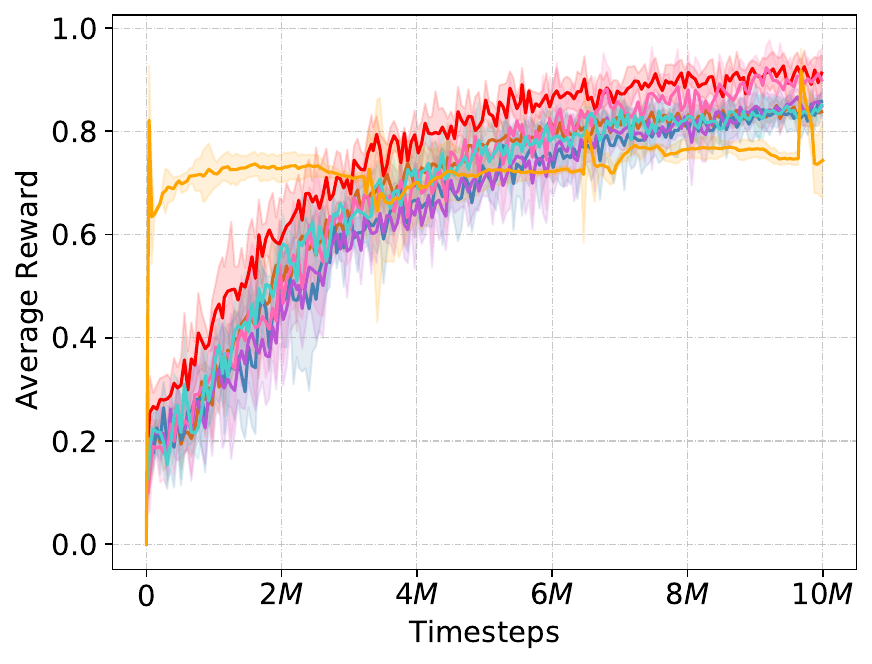}}
	\subfloat[\small SynthSeq]{\includegraphics[width = 0.314\textwidth]{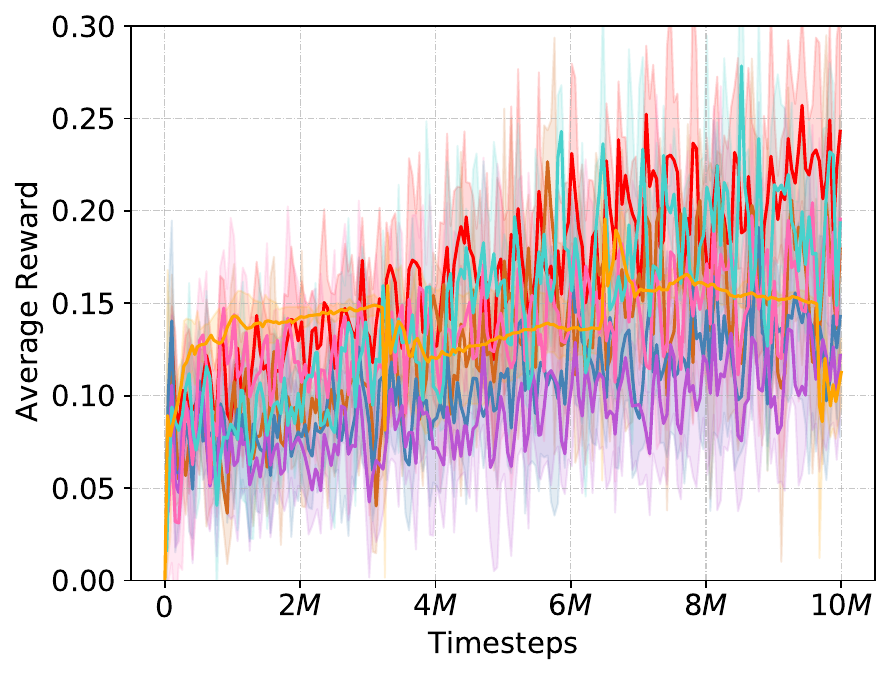}}
	\hfill
	
	\subfloat[\small BossLevel]{\includegraphics[width = 0.314\textwidth]{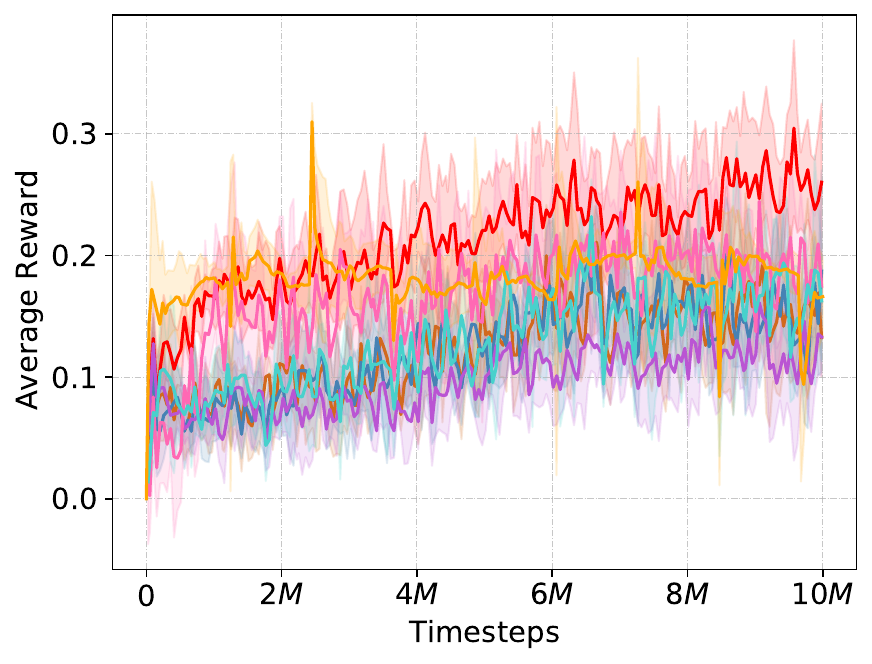}}
	\subfloat[\small BossLevelNoUnlock]{\includegraphics[width = 0.314\textwidth]{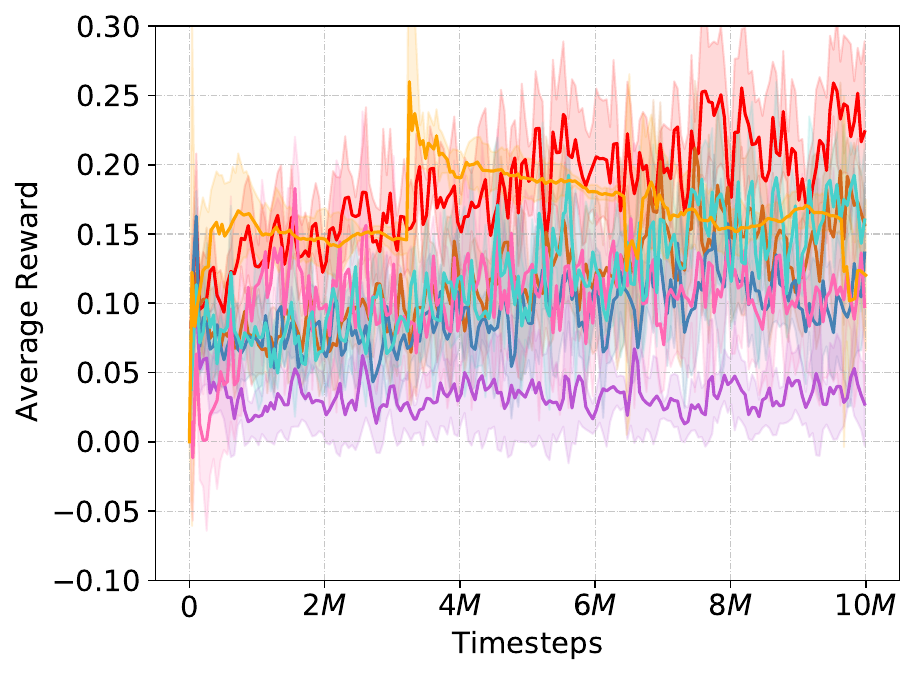}}
	\subfloat[\small MiniBossLevel]{\includegraphics[width = 0.31\textwidth]{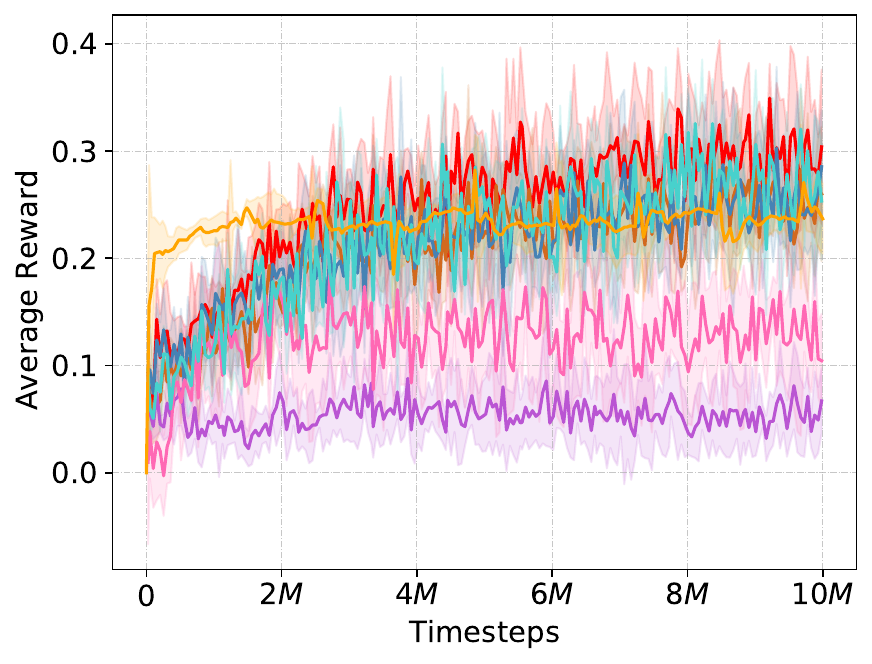}}
\caption{Training curve comparisons on six complicated tasks, averaged over 10 independent running seeds.}
\label{fig:exp-2-6}
\end{figure*}

\subsection{Results on Training Curves}

We evaluate our method on multiple difficulty levels of BabyAI, which require mastering various competencies, including navigation, location recognition, sequence understanding, object manipulation, and more. For example, \textit{GotoSeq} only requires navigating a room or maze, recognizing the location, and understanding the sequence subtasks while \textit{BossLevel} requires mastering all competencies including navigating, opening, picking up, putting next, and so on.
We conduct experiments on these levels to test the effectiveness of our algorithm in complicated and dynamic environments, as elaborated in Appendix~\ref{app:more-complcated}.


\revise{
Here, we report the average episode reward by training steps for each level, averaged over 10 independent running seeds to plot the reward curve with standard deviation.
In \textit{GoToSeqS5R2} which is the relatively simplest level of the six, all the test rewards keep increasing as the training steps increase and finally almost reach the best.
It is worth noting that although LEAP had a relatively good starting point in its early stages, the upwards trend was not evident in the later stages. We speculate that this is because LEAP, during its training, had encountered sufficient historical data, giving the model a higher starting point. However, the model's capacity in the later stages seemed relatively weaker, preventing it from achieving even higher performance.
In other complicated tasks, our algorithm consistently obtains higher rewards than all the baselines in the different levels of BabyAI shown in Fig.~\ref{fig:exp-2-6}, while the popular SOTA algorithms, such as PPO and SAC almost fail. This indicates that our method is capable of macroplanning and robust enough to handle complicated tasks by decomposing the overall process into a sequence of execution subtasks.

Although we only trained 10M timesteps due to the limitation of computing resources, our method has a higher starting point than the baseline model, and it has been rising. However, the gain of the baselines at the same timestep is not obvious in most cases.
This significant improvement is mainly due to two aspects.
On the one hand, the adequate knowledge representation facilitates the inference capacity of our model and provides a high-level starting point for warming up the training.
On the other hand, the learned subtask planning strategy can make rational decisions while confronting various complicated situations.
}

\revise{
\subsection{Results on Execution Performance}}

\revise{
To further test the execution performance of all methods, we freeze the trained policies and execute for 100 episodes, and report the mean and standard deviations in Table~\ref{tab:exection}.
Our algorithm can consistently and stably outperform all the competitive baselines by a large margin, which demonstrates the benefits of our framework.
}

\begin{table}[!h]
    \centering
    \caption{\revise{Execution performance on various complicated tasks. We report the average reward with standard deviations of 100 testing episodes with best values in bold.}}
    \begin{tabular}{m{2.3cm}<{\centering} m{1.1cm}<{\centering} m{0.9cm}<{\centering} m{0.9cm}<{\centering} m{0.9cm}<{\centering} m{0.9cm}<{\centering} m{0.9cm}<{\centering} m{0.9cm}<{\centering}} 
    \toprule
        ~ & \footnotesize \revise{Ours} & \footnotesize \revise{CORRO} & \footnotesize \revise{PPO} & \footnotesize \revise{SAC} & \footnotesize \revise{SGT} & \footnotesize \revise{DGRL} & \footnotesize \revise{LEAP}    \\ \hline  \vspace{0.3em}
        \footnotesize  \revise{GoToSeq} &   
        \revise{\footnotesize \textbf{0.30}\tiny (0.07)} &  
        \revise{\footnotesize 0.23\tiny (0.06)} &  
        \revise{\footnotesize 0.14\tiny (0.04)} &   
        \revise{\footnotesize 0.05\tiny (0.04)} &    
        \revise{\footnotesize 0.15\tiny (0.08)} &
        \revise{\footnotesize 0.19\tiny (0.07)} &    
        \revise{\footnotesize 0.17\tiny (0.04)} 
        \\ \vspace{0.3em}
        \footnotesize \revise{GoToSeqS5R2} & 
        \revise{\footnotesize \textbf{0.91}\tiny (0.02)}& 
        \revise{\footnotesize {0.84}\tiny (0.02)} & 
        \revise{\footnotesize 0.84\tiny (0.03)} &  
        \revise{\footnotesize 0.85\tiny (0.02)} &
        \revise{\footnotesize \textbf{0.91}\tiny (0.04)} & 
        \revise{\footnotesize 0.67\tiny (0.19)} &
        \revise{\footnotesize {0.73}\tiny (0.06)}  
        \\  \vspace{0.3em}
        \footnotesize  \revise{SynthSeq} & 
        \revise{\footnotesize \textbf{0.22}\tiny (0.05)} & 
        \revise{\footnotesize 0.16\tiny (0.05)}& 
        \revise{\footnotesize 0.12\tiny (0.05)}&  
        \revise{\footnotesize 0.12\tiny (0.04)}& 
        \revise{\footnotesize 0.18\tiny (0.05)} &
        \revise{\footnotesize 0.15\tiny (0.06)} &
        \revise{\footnotesize {0.14}\tiny (0.03)}  
        \\ \vspace{0.3em}
        \footnotesize  \revise{BossLevel} & 
        \revise{\footnotesize \textbf{0.26}\tiny (0.05)} & 
        \revise{\footnotesize 0.15\tiny (0.04)} &  
        \revise{\footnotesize 0.14\tiny (0.05)} &  
        \revise{\footnotesize 0.12\tiny (0.04)} &
        \revise{\footnotesize 0.19\tiny (0.06)} &
        \revise{\footnotesize 0.13\tiny (0.05)} &
        \revise{\footnotesize {0.18}\tiny (0.04)}  
        \\ \vspace{0.3em}
        \footnotesize \revise{BossLevelNoUnlock} & 
        \revise{\footnotesize \textbf{0.24}\tiny (0.04)} & 
        \revise{\footnotesize 0.17\tiny (0.04)} & 
        \revise{\footnotesize 0.10\tiny (0.05)} &  
        \revise{\footnotesize 0.04\tiny (0.03)} &  
        \revise{\footnotesize 0.11\tiny (0.05)} &
        \revise{\footnotesize 0.12\tiny (0.05)} &
        \revise{\footnotesize {0.16}\tiny (0.03)}  
        \\ \vspace{0.3em}
        \footnotesize  \revise{MiniBossLevel} & 
        \revise{\footnotesize \textbf{0.29}\tiny (0.06)} & 
        \revise{\footnotesize 0.26\tiny (0.05)} &  
        \revise{\footnotesize 0.25\tiny (0.05)} &  
        \revise{\footnotesize 0.06\tiny (0.03)}  &
        \revise{\footnotesize 0.14\tiny (0.06)} &
        \revise{\footnotesize 0.22\tiny (0.07)} &
        \revise{\footnotesize {0.23}\tiny (0.03)}  
        \\ 
        \bottomrule
    \end{tabular}
    \label{tab:exection}
\end{table}

\subsection{Results on Various Tree Widths}

\begin{figure}[ht!]
\centering
	\subfloat[\small BossLevel]{\includegraphics[width = 0.3\textwidth]{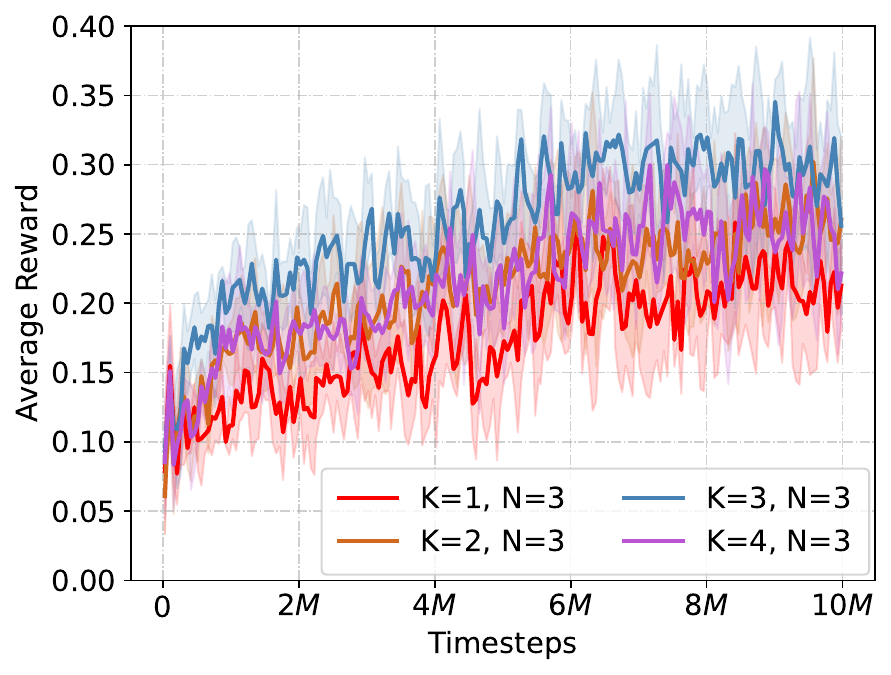}}
    \subfloat[\small {BossLevelNoUnlock}]{\includegraphics[width = 0.3\textwidth]{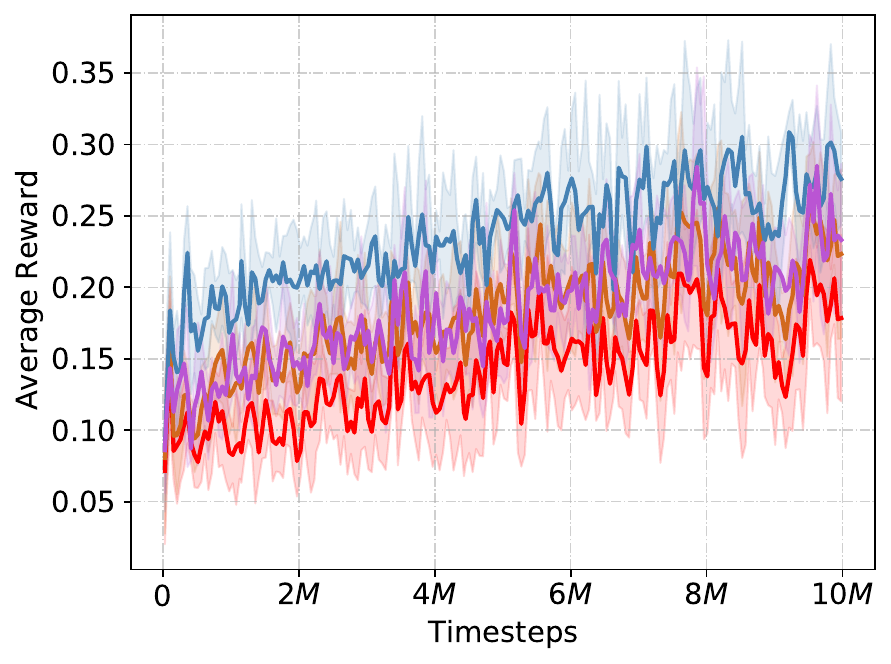}
	\label{app:BossLevelNoUnlock-exp3-1}
	}
	\subfloat[\small {MiniBossLevel}]{\includegraphics[width = 0.3\textwidth]{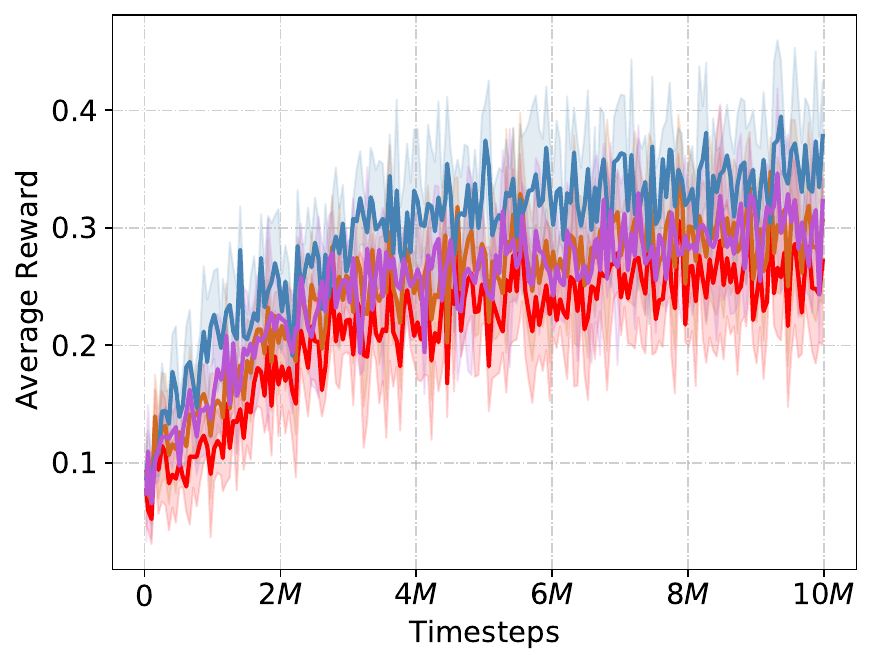}
	\label{app:MiniBossLevel-exp3-1}
	}
	\hfill
 
	\subfloat[\small {GotoSeq}]{\includegraphics[width = 0.3\textwidth]{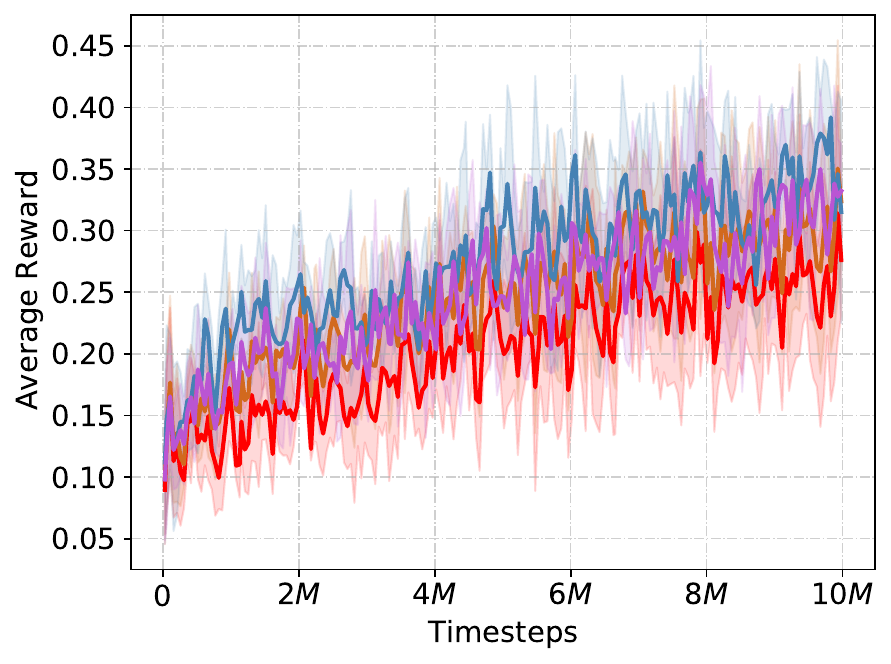}}
     \subfloat[\small {GoToSeqS5R2}]{\includegraphics[width = 0.3\textwidth]{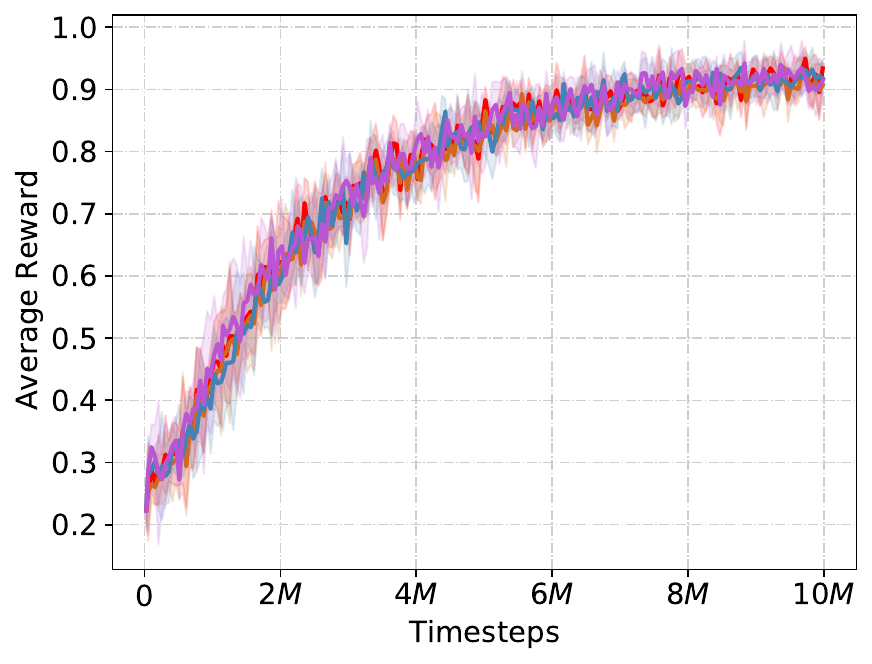}
    	\label{app:GoToSeqS5R2-exp3-1}
    	}
	\subfloat[\small {SynthSeq}]{\includegraphics[width = 0.3\textwidth]{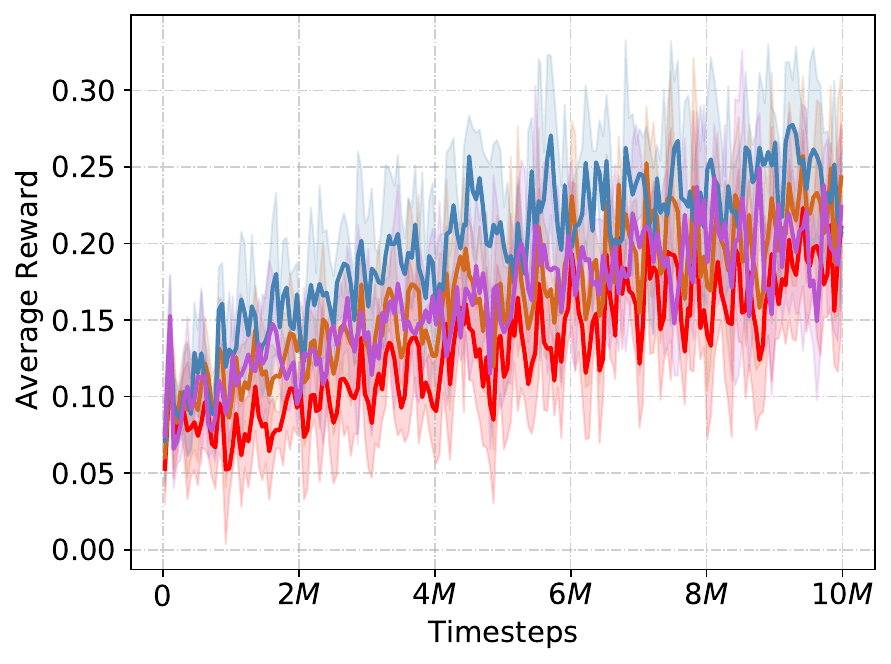}}
	
\caption{Sensitivity analyses on various tree widths.}
\label{fig:exp-3-width}
\end{figure}

After evaluating the training and execution performance of our technique, we performed sensitivity analyses on the width of the planning tree. The results are shown in Fig.~\ref{fig:exp-3-width}, where we kept the tree depth at $3$, as in the main experiments. 

The horizontal axis represents the training steps, and the vertical axis shows the average episode reward with a standard deviation, averaged across 10 independent running seeds. Our intuition suggests that a wider tree would result in a more diverse subtask distribution. However, we found that the gain is not the largest when the treewidth is the largest.
We speculate that a larger tree width provides more information but weakens the pertinence of problem-solving, leading to confusion. Hence, it is critical to select appropriate treewidth. In the future, we plan to conduct an adaptive selection of the optimal tree width.

\subsection{Results on Various Tree Depths}
\begin{figure}[ht!]
\centering
	\subfloat[\small BossLevel]{\includegraphics[width = 0.3\textwidth]{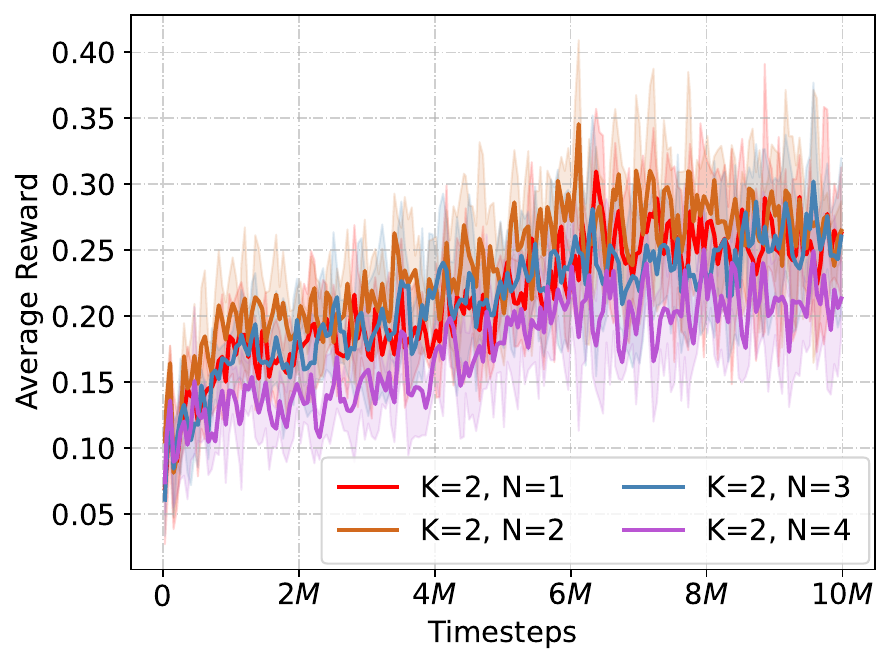}}
    \subfloat[\small {BossLevelNoUnlock}]{\includegraphics[width = 0.3\textwidth]{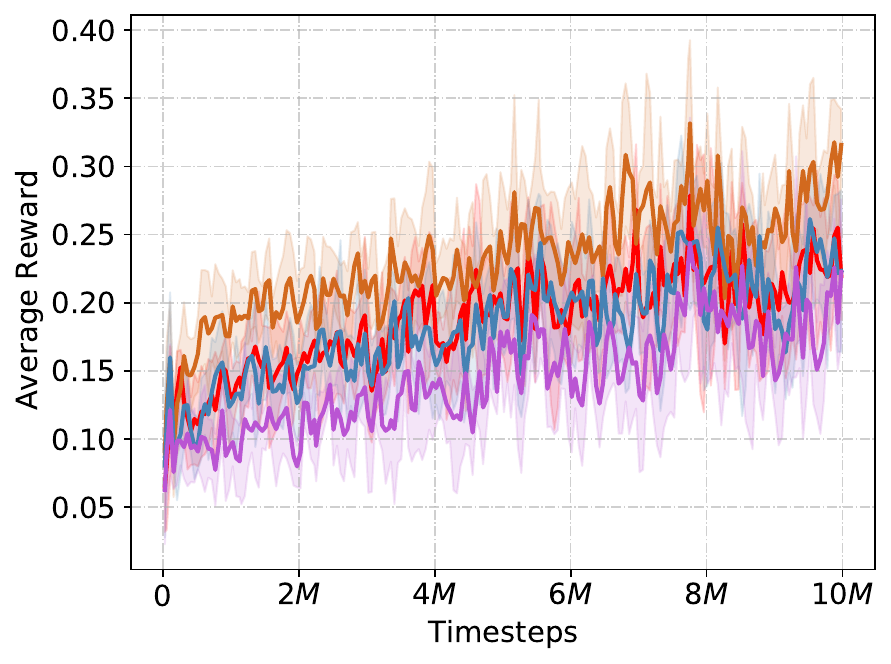}
	\label{app:BossLevelNoUnlock-exp3-2}
	}
	\subfloat[\small {MiniBossLevel}]{\includegraphics[width = 0.3\textwidth]{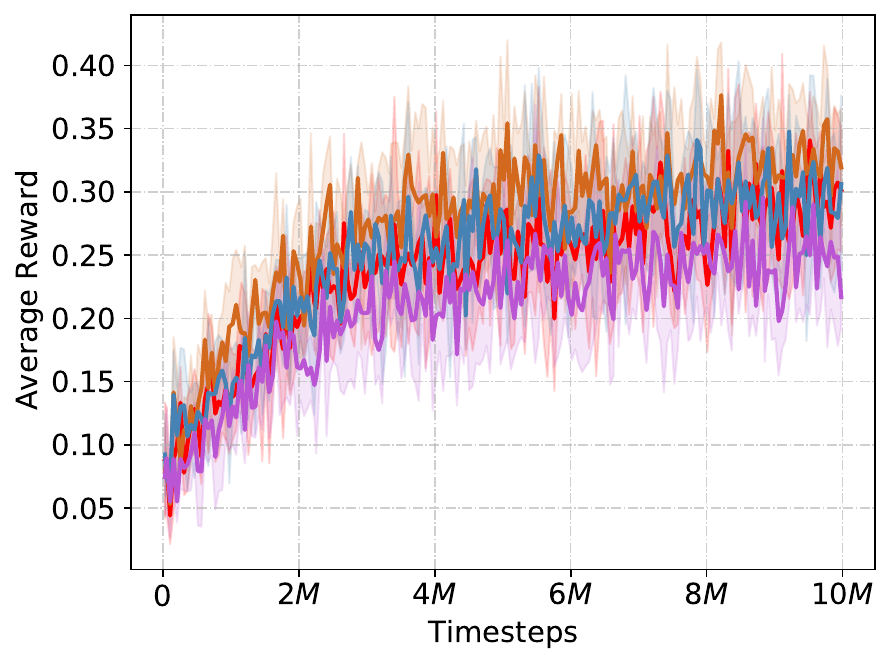}
	\label{app:MiniBossLevel-exp3-2}
	}
	\hfill
 
	\subfloat[\small {GotoSeq}]{\includegraphics[width = 0.3\textwidth]{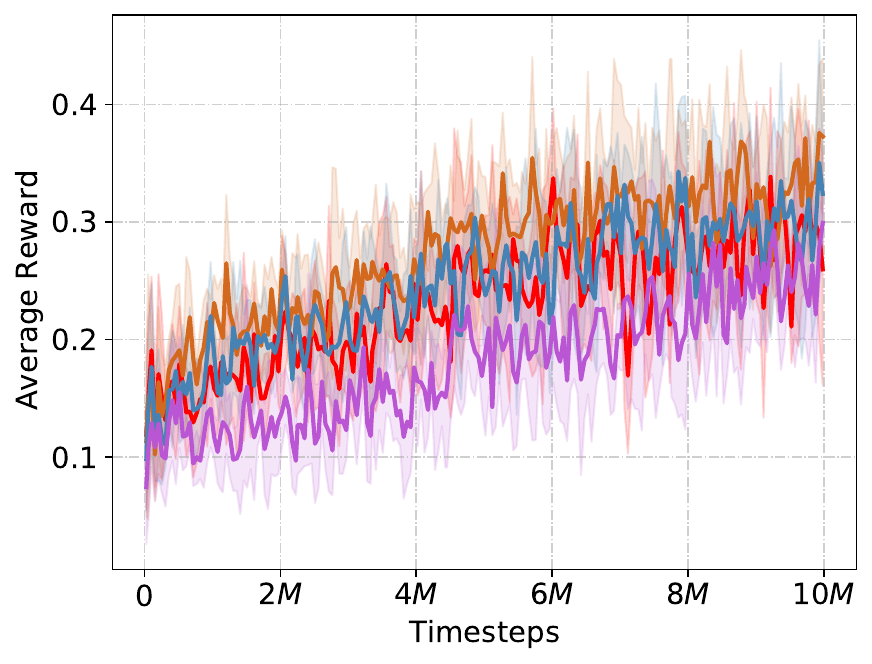}}
    \subfloat[\small {GoToSeqS5R2}]{\includegraphics[width = 0.3\textwidth]{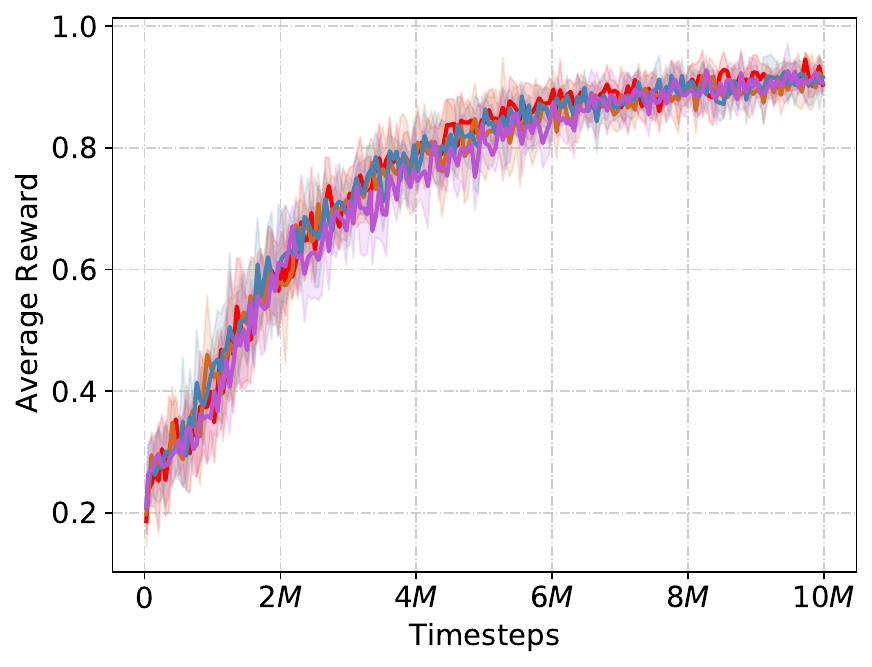}
	\label{app:GoToSeqS5R2-exp3-2}
	}
	\subfloat[\small {SynthSeq}]{\includegraphics[width = 0.3\textwidth]{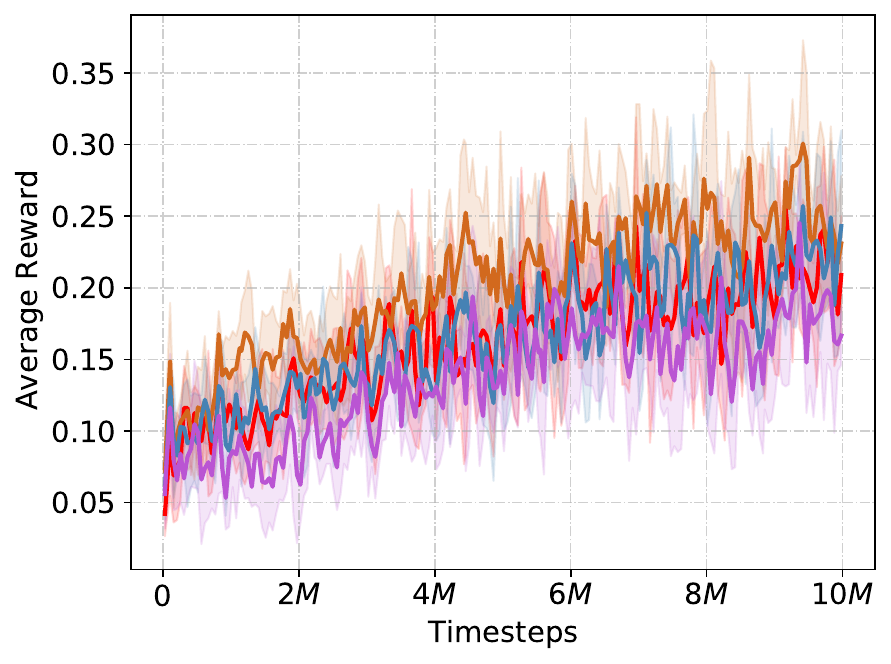}}
	
\caption{Sensitivity analyses on various tree depths.}
\label{fig:exp-3-depth}
\end{figure}


In Fig.~\ref{fig:exp-3-depth}, we present additional sensitivity analyses to investigate the impact of various tree depths on the behavior of our planning tree. The treewidth is fixed as $2$, the same as the main experiments.
The depth represents the model's capacity to predict the future of the planning tree. Our results show that when the depth is set to $N=2$ or $N=3$, our model achieves comparable performance to the main experiments. However, the curves for $N=4$ show degraded performance compared to the other depths.
We speculate that the accumulation of prediction errors about the future at each step can lead the model to make suboptimal decisions, despite the improved foresight ability in making global decisions.
Hence, selecting an appropriate depth for the planning tree is crucial for our model's success, and it is a potential future direction to explore adaptive selection techniques.

\subsection{Results on Subtask Execution Statistics}

\begin{figure}[h!]
    \centering
\includegraphics[width=0.66\linewidth]{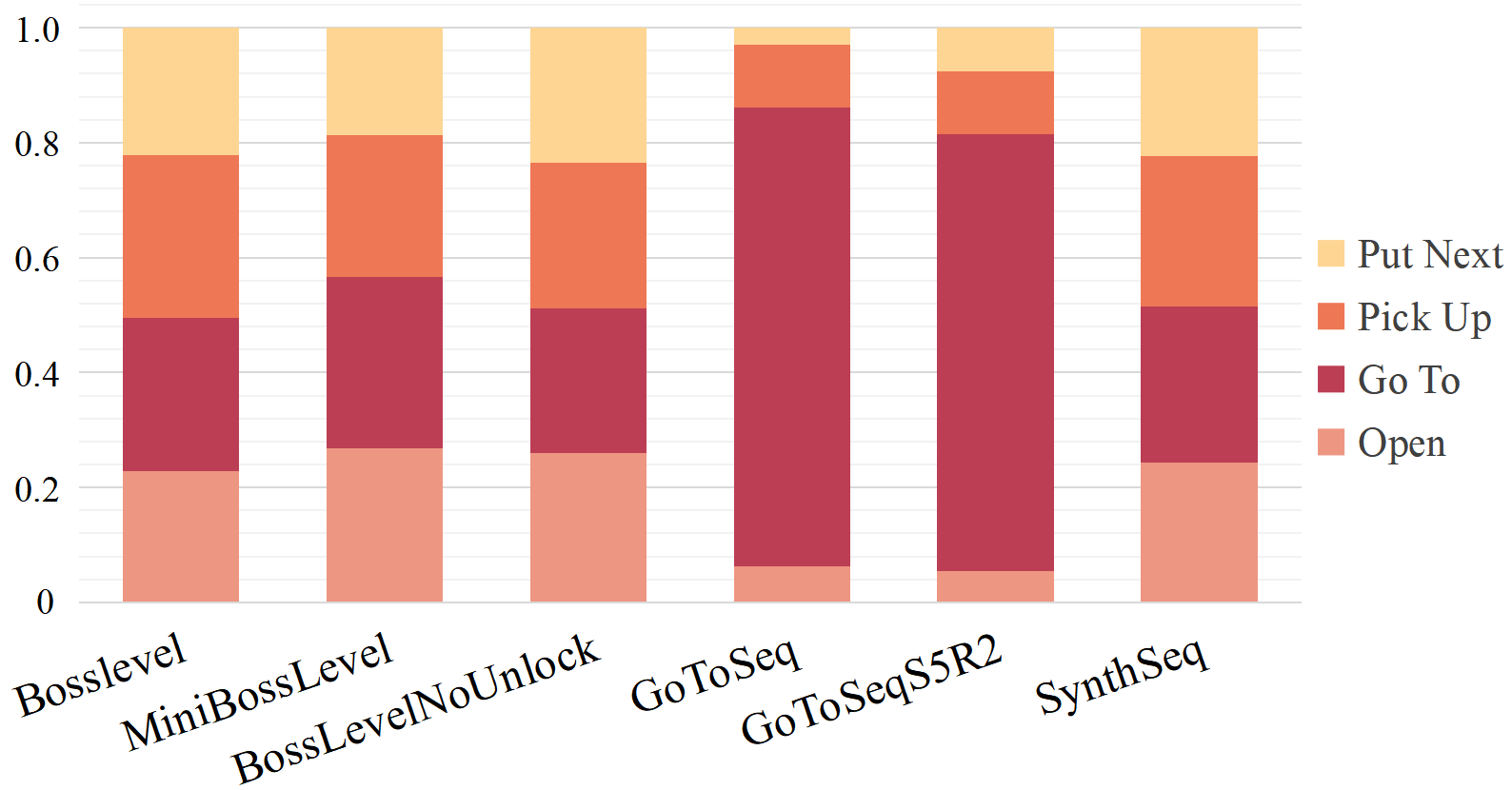}
\caption{The bars represent the mean proportion of each subtask executed by our trained model over 1000 episodes.}
\label{fig:sub-exe}
\end{figure}

In Fig.~\ref{fig:sub-exe}, we present the subtask proportions on six complex tasks after 1000 episodes of testing. The observed subtask distribution is highly consistent with the predefined task missions, validating the rationality of our approach. Our model focuses on executing the necessary sequences related to the ``Go To'' task in \textit{GoToSeq} and \textit{GoToSeqS5R2} for most cases. However, occasionally, other subtasks are selected, which we attribute to the inaccuracies of forward prediction while expanding the tree depth. For the remaining four comprehensive tasks, our model executes subtasks at roughly equal frequencies.

%% file: 6-con.tex
\section{Conclusion}


In this paper, we have considered the difficulties of task planning in solving complicated problems, and propose a two-stage planning framework. The multiple-encoder and individual-predictor regime can extract discriminative subtask representations with dynamic knowledge from priors.
The top-k subtask planning tree, which customizes subtask execution plans to guide decisions, is important in complicated tasks.
Generalizing our model to unseen and complicated tasks demonstrates significant performance and exhibits greater potential to deploy it to real-life complicated tasks.
Overall, our approach provides a promising solution to the task planning problem in complicated scenarios.
Future research directions could include automatic learning of explainable subtasks and incorporating global planning as a reference for human decision-making in real-life applications.

%% file: 7-ack.tex
\section*{Acknowledgement}

This work was supported by the National Key R\&D Program of China
(No.2022ZD0116405), the Strategic Priority Research
Program of the Chinese Academy of Sciences (No.XDA27030300), and the National Natural Science Foundation of China under Grant 62073324.

%% file: 7-append.tex

\clearpage

\begin{appendices}




\section{Detailed Description about the Intuition for d-UCB}
\label{app:d-uct}

\begin{figure}[h!]
    \centering
\includegraphics[width=0.7\linewidth]{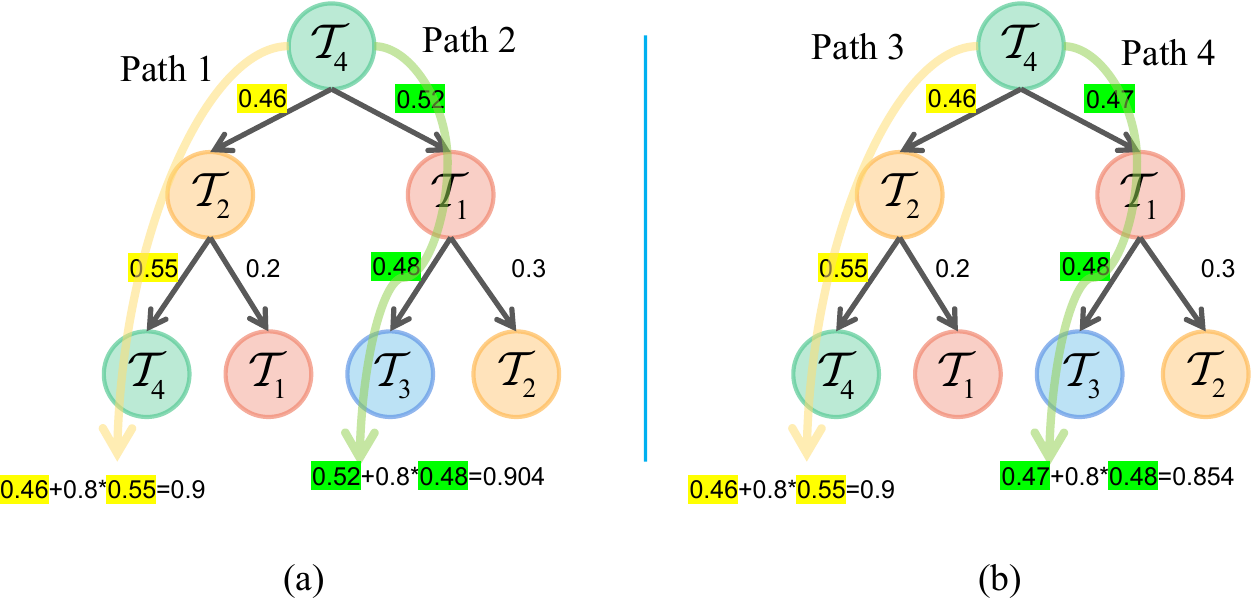}
\caption{Two instance trees are visualized to verify the effect of d-UCB described in Section~\ref{sec:d-ucb}. To ignore the effect of the UCB value, we set the value of all nodes as $1$.}
\label{fig:d-uct}
\end{figure}
\vspace{-10pt}

Here, we provide another two methods.
One is called max selection (MS), which means that MS will pick the path in that the cumulative probability is maximum. Next, we will compare it with our d-UCB method.
Another is called greedy selection (GS), which means that GS will pick the current maximum probability layer by layer.
Next, we will compare it with our d-UCB method.

As shown in Fig.~\ref{fig:d-uct}(a), MS will select path $1$, which obviously places a greater weight on the future with greater uncertainty.
However, our d-UCB will choose path $2$, which tends to maximize the immediate benefit by discounting the future uncertain score, which is consistent with our intuition.

Moreover, as shown in Fig. ~\ref{fig:d-uct}(b), unlike MS tending to pick the edge with the highest probability, our d-UCB will choose path $3$.
Our d-UCB also considers the optimal subtask to perform in the future when the nearest paths exhibit similar transition probabilities.

\section{Detailed Description about Complicated Tasks}
\label{app:more-complcated}
\begin{figure}[ht!]
\centering
	\subfloat[\scriptsize ‘BossLevel-v0’]{\includegraphics[width = 0.3\textwidth]{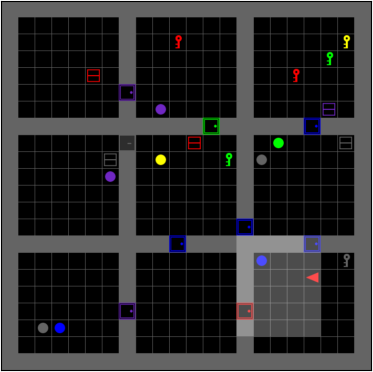}
	\label{app:BossLevel}
	}
	\subfloat[\scriptsize ‘BossLevelNoUnlock-v0’]{\includegraphics[width = 0.3\textwidth]{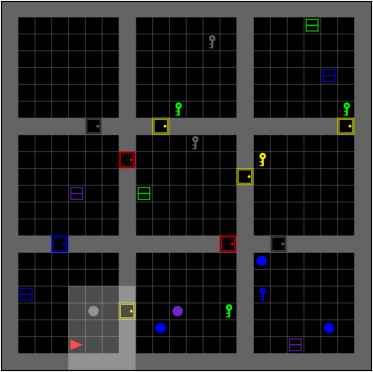}
	\label{app:BossLevelNoUnlock}
	}
	\subfloat[\scriptsize ‘MiniBossLevel-v0’]{\includegraphics[width = 0.3\textwidth]{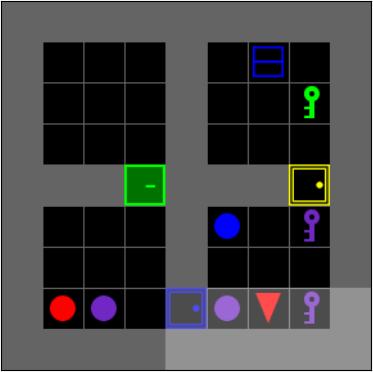}
	\label{app:MiniBossLevel}
	}
	\hfill
	\subfloat[\scriptsize ‘GoToSeq-v0’]{\includegraphics[width = 0.3\textwidth]{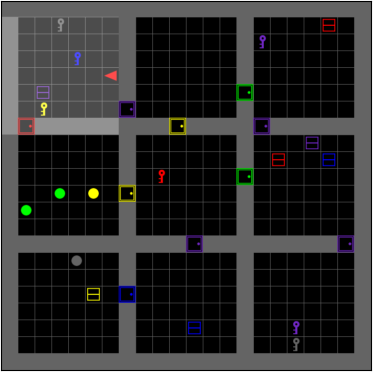}
	\label{app:GoToSeq}
	}
	\subfloat[\scriptsize ‘GoToSeqS5R2-v0’]{\includegraphics[width = 0.3\textwidth]{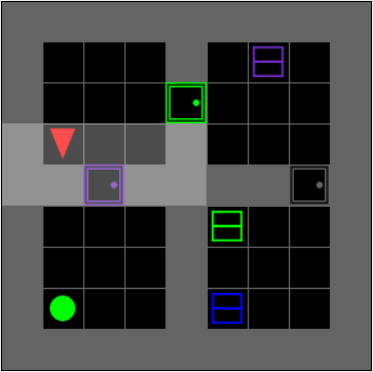}
	\label{app:GoToSeqS5R2}
	}
	\subfloat[\scriptsize ‘SynthSeq-v0’]{\includegraphics[width = 0.3\textwidth]{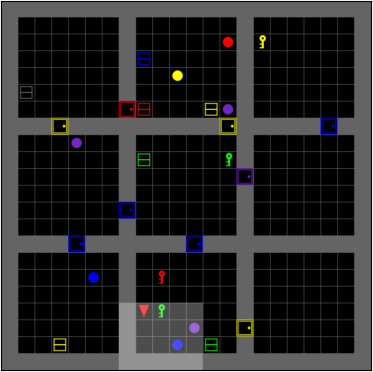}
	\label{app:SynthSeq}
	}
	
\caption{The diagrams for six complicated tasks in BabyAI.}
\label{fig:exp-4}
\end{figure}
\vspace{-10pt}

{\textbf{BossLevel.}}The level requires the agent to implement any union of instructions produced according to Baby Language grammar, it involves the content of all other levels. For example, ``open a door and pick up the green key after you go to a ball and put a red box next to the red ball'' is a command of the level. This level requires all the competencies including navigating, unblocking, unlocking, opening, going to, picking up, putting next, and so on.


{\textbf{BossLevelNoUnlock.}}The doors of all rooms are unlocked which the agent can go through without unlocking by a corresponding key. For example, the command ``open a purple door, then go to the yellow box and put a yellow key next to a box'' does not need the agent to get a key before opening a door. This level requires all the competencies such as BossLevel except for unlocking.


{\textbf{MiniBossLevel.}}The level is a mini-environment of BossLevel with fewer rooms and a smaller size. This level requires all the competencies such as BossLevel except unlocking.


{\textbf{GoToSeq.}}The agent needs to complete a series of go-to-object commands in sequence. For example, the command "go to a grey key and go to a purple key, then go to the green key" only includes go-to instructions. GotoSeq only requires
navigating a room or maze, recognizing the location, and understanding the sequence subtasks.


{\textbf{GoToSeqS5R2.}}
The missions are just similar to GoToSeq, which requires the only competency of ``go to''. Here, S5 and R2 refer to the size and the rows of the room, respectively.



{\textbf{SynthSeq.}}The commands, which combine different kinds of instructions, must be executed in order. For example, "Open the purple door and pick up the blue ball, then put the grey box next to a ball".
This level requires competencies such as GotoSeq except inferring whether to unlock or not.

\section{Network Architecture}
\label{app:para-nn}
There is a summary of all the neural networks used in our framework regarding the network structure, layers, and activation functions.

\begin{sidewaystable}
\sidewaystablefn%
\begin{center}
\begin{minipage}{\textheight}
\caption{Summary of the Network Architecture}\label{tab:net}
\begin{tabular*}{\textheight}{@{\extracolsep{\fill}}ccccc@{\extracolsep{\fill}}}
       \toprule
       ~ & Network Structure &  Layers & Hidden Size & Activation Functions  \\
       \midrule
       Subtask Encoder & MLP & 6 &   [256, 256, 128, 128, 32]  & ReLu \\
       Shared Subtask Predictor & MLP & 3 &  [64, 128]  & ReLu   \\
       Query Encoder &   MLP & 4 & [32, 16, 8] & ReLu \\
       Tree Node Generation &  CausalSelfAttention & 1 & - & Softmax \\
       Feature Extractor & CNN+FiLMedBlock & - & - & ReLu \\
       Policy Actor  &   MLP & 2 & 64 &Tanh \\
       Policy Critic & MLP & 2 & 64 & Tanh\\
       \botrule
    \end{tabular*}

\footnotetext{Please note that '-' denotes that we refer readers to check the open source repository about FiLMedBlock and CausalSelfAttention, see \href{https://github.com/caffeinism/FiLM-pytorch}{https://github.com/caffeinism/FiLM-pytorch} and \href{https://github.com/sachiel321/Efficient-Spatio-Temporal-Transformer}{https://github.com/sachiel321/Efficient-Spatio-Temporal-Transformer}, respectively.}
\end{minipage}
\end{center}
\end{sidewaystable}

\end{appendices}